\documentclass{article}
\usepackage{etoolbox}       %
\usepackage{tabularx}       %

\usepackage{graphicx}  %

\usepackage[parfill]{parskip}
\usepackage{amsmath}
\usepackage{amsthm}
\usepackage{mathtools}  %
\usepackage[dvipsnames]{xcolor}         %
\usepackage{float}  %
\usepackage{listings}  %

\newtoggle{arxiv}
\toggletrue{arxiv}

  \usepackage[parfill]{parskip}
  \usepackage[citestyle=authoryear-comp,sorting=nyt,maxbibnames=99,backend=biber]{biblatex}
  \newcommand{\citep}{\parencite}
  \newcommand{\citet}{\textcite}
  \newlength{\defbaselineskip}
  \RequirePackage[T1]{fontenc}
  \RequirePackage[tt=false, type1=true]{libertine}
  \RequirePackage[varqu]{zi4}
  \RequirePackage[libertine]{newtxmath}

\usepackage{bm}

\usepackage[inline]{enumitem}
\usepackage{booktabs}
\usepackage[dvipsnames]{xcolor}  %
\usepackage{subcaption}
\usepackage{multirow}
\usepackage{wrapfig}
\usepackage{algorithm,algorithmicx,algpseudocode}
\usepackage{nicematrix}

\usepackage[frozencache,cachedir=.]{minted}

\usepackage[colorlinks=true,linkcolor=blue,citecolor=magenta,urlcolor=red]{hyperref}
\usepackage[capitalise,noabbrev]{cleveref}

\usepackage{import}

\usepackage{pifont}

  \title{Wonderful Matrices: Combining for a More Efficient and Effective \\ Foundation Model Architecture}
  \usepackage{authblk}
  \author[$^1$]{Jingze Shi\thanks{This author contributed Algorithm design and Experiment verification.}}
  \author[$^2$]{Bingheng Wu\thanks{This author contributed Data analysis and Datasets processing.}}
  \affil[ ]{Independent Researcher}
  \affil[ ]{{\texttt{losercheems@gmail.com}}, {\texttt{wubingheng52136@gmail.com}}}
  \date{}

\begin{document}

  \maketitle

\begin{abstract}
In order to make the foundation model more efficient and effective, our idea is combining sequence transformation and state transformation.
First, we prove the availability of rotary position embedding in the state space duality algorithm, which reduces the perplexity of the hybrid quadratic causal self-attention and state space duality by more than 4\%, to ensure that the combining sequence transformation unifies position encoding.
Second, we propose dynamic mask attention, which maintains 100\% accuracy in the more challenging multi-query associative recall task, improving by more than 150\% compared to quadratic causal self-attention and state space duality, to ensure that the combining sequence transformation selectively filters relevant information.
Third, we design cross domain mixture of experts, which makes the computational speed of expert retrieval with more than 1024 experts 8 to 10 times faster than the mixture of experts, to ensure that the combining state transformation quickly retrieval mixture.
Finally, we summarize these matrix algorithms that can form the foundation model: Wonderful Matrices, which can be a competitor to popular model architectures.
\end{abstract}

\section{Introduction}
\label{sec:introduction}

The backbone of modern foundation models usually consists of two main parts: one is sequence transformation, which assigns dependencies to elements; the other is state transformation, which assigns knowledge to elements.

In the sequence transformation part, efficient algorithms aim to compress element dependency information in a limited state, while effective sequence transformation algorithms aim to store all element dependencies.
Transformer~\citep{vaswani2017attention} Architecture is popular in modern language modeling, it directly captures the relationship between any two elements in the sequence by calculating the causal mask matrix, which can effectively handle long-distance dependency problems. However, the architecture has a major drawback: the quadratic complexity of the Quadratic Causal Self-Attention in the sequence transformation part limits the ability to handle long contexts.
State Space Model~\citep{dao2024ssd} Architecture came into being, it balances the quadratic and linear calculation methods of relevant elements by calculating the semiseparable matrix, which can achieve linear scaling of sequence length during training and maintain a constant state size during generation. However, the architecture also has a major drawback: the dependency state of the State Space Duality in the sequence transformation part does not expand with the sequence length to cause dependency bias.

In the state transformation part, efficient algorithms aim to sparsely activate knowledge parameters related to elements, while effective algorithms aim to densely activate knowledge parameters related to elements.
Gated Multi-Layer Perceptron~\citep{glu2020} consists of a Linear layer with dense activation and an activation function, it controls the flow of information through gate units, which can suppress the output of certain neurons. However, the structure has a major drawback: each output unit of the Linear layer receives information from all input units, causing the computational complexity to increase with the number of units, leading to difficult to expand.
Then a series of sparse mixture of experts structures appeared, among which the most efficient is the Mixture of A Million Experts~\citep{he2024moame} mainly composed of embedding layers and activation functions, which maintains computational efficiency through parameter-efficient expert retrieval. However, the structure also has a major drawback: one input unit of the Embedding layer only activates one output unit, causing the sharing ratio to not increase with the number of units, 
leading to redundancy to stored.

To build a model that is both efficient and effective, the key is to balance the combination relationship between different sequence transformations and state transformations. Our main goal is to integrate the State Space Duality algorithm with the Quadratic Causal Self-Attention algorithm, combining the Linear layer with dense activation and the Embedding layer with sparse activation to overcome their respective limitations. Although this hybrid algorithm foundation model architecture will lose some of the extreme excellence of specific tasks, it will have more comprehensive capabilities.

\begin{figure}[t]
    \centering
    \includegraphics[width=\linewidth]{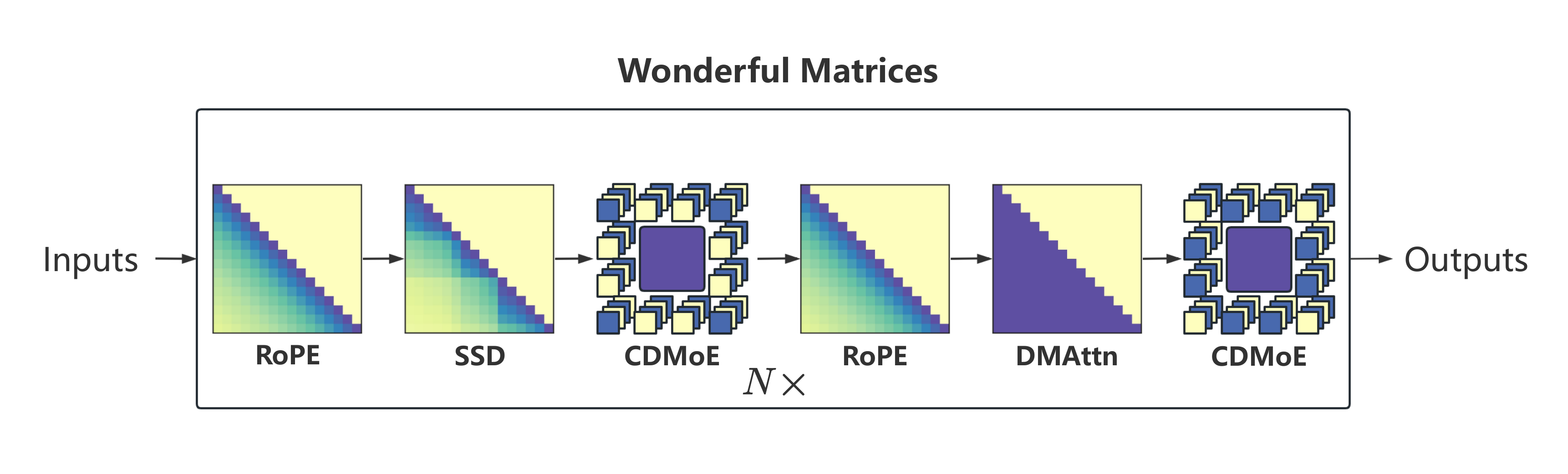}
    \caption{
      \textbf{Wonderful Matrices Architecture}.
      Shows the matrices used in the Wonderful Matrices Architecture, including the Rotary Position Embedding Matrix, State Space Duality Matrix, Dynamic Mask Attention Matrix, Cross Domain Mixture of Experts Matrix, and the process of using these matrices.
    }
    \label{fig:wonderful_matrices}
\end{figure}

\paragraph{Position Encoding.}
The key to combining the State Space Duality algorithm and the Quadratic Causal Self-Attention algorithm is to integrate the position information.
In Mamba~\citep{gu2023mamba}, the position information is implicitly provided by causal convolution, and then the matrix D skip connect the input and output of the state space algorithm to re-extend the discrete position information.
In Mamba2~\citep{dao2024ssd}, it is mentioned that the cumulative product of the gate can be directly used to allow two positions to interact with each other as the position information of the state space algorithm.
However, convolution operations for position encoding are time-consuming, and recursive position encoding can only be applied to State Space Duality and cannot be applied to Quadratic Causal Self-Attention.
Therefore, we prove the availability of Rotary Position Embedding~\citep{su2021roformer} in State Space Duality to unify the position encoding.

\paragraph{Selective Transformation.}
Another key to combining the State Space Duality algorithm with the Quadratic Causal Self-Attention algorithm is the same transformation state.
In the State Space Duality algorithm, selective filtering of sequence state information is achieved through a gate matrix, which is equivalent to a trainable dynamic mask.
In the Quadratic Causal Self-Attention algorithm, future information leakage is prevented by a predefined causal mask, which is a static mask that relies entirely on human design for additional information filtering.
Therefore, we propose dynamic mask attention to allow Quadratic Causal Self-Attention to dynamically adjust attention score masks based on the current value state, to match the selectivity of the State Space Duality algorithm.

\paragraph{Full Utilization.}
The key to combining the Linear layer with the Embedding layer is to full utilize the parameters.
Knowledge is widely distributed in different domains, which are interconnected through common general knowledge and cross domain knowledge.
A feedforward network composed solely of Linear layers or Embedding layers cannot fully utilize the parameters to store this knowledge.
Therefore, we design Cross Domain Mixture of Experts, which has shared parameters for storing general knowledge and professional parameters for storing domain specific knowledge, and can significantly improve the granularity of experts without causing a rapid decrease in computational speed.

\paragraph{Architecture Design.}
We use the Rotary Position Embedding Matrix as the position encoding method, the State Space Duality Matrix and the Dynamic Mask Attention Matrix as the sequence transformation, and the Cross Domain Mixture of Experts Matrix as the state transformation. These matrices form the Wonderful Matrices architecture~\ref{fig:wonderful_matrices}.

We evaluate the Wonderful Matrices architecture on the language modeling task, including the each module and the overall architecture. 
The code is open-sourced at \url{https://github.com/LoserCheems/WonderfulMatrices}.

\section{Related Work}
\label{sec:related_work}

\paragraph{Quadratic Causal Self-Attention.}
Self-Attention is a mechanism that computes the relevance scores between each element in the sequence and all other elements, allowing each element to "attend" to other elements. The most important variant of attention is the Quadratic Self-Attention.
A notable feature of Quadratic Self-Attention is that it can capture dependencies between any positions in the input sequence, without being limited by distance, and the state expands with the sequence length, which gives it an advantage in capturing long-range dependencies in long sequences.
In causal language modeling, a causal mask matrix is usually added to the attention score matrix to prevent future information leakage. We refer to it as Quadratic Causal Self-Attention.

\begin{align*}
Y &= \operatorname*{softmax}({QK^\top}) \cdot V
\end{align*}

\paragraph{State Space Duality.}
Many variants of Quadratic Causal Self-Attention are proposed based on the calculation improvement of the attention score matrix. The most important variant is linear attention~\citep{katharopoulos2020transformers}, which rewrites $(QK^\top) \cdot V = Q \cdot (K^\top V)$ by folding softmax into the kernel feature map and using the kernel properties of matrix multiplication. In the case of causal attention, they show that when the causal mask is merged to the left $(L \circ QK^\top) \cdot V$, where $L$ is a lower triangular matrix, the right side can be expanded into a recursive form, allowing attention to perform linear autoregressive reasoning.
In Transformers are SSMs~\citep{dao2024ssd}, the State Space Duality is used to prove that by implementing the semiseparable matrix $M = L \circ CB^\top = L \circ QK^\top$ and performing quadratic matrix-vector multiplication, the result is equivalent to quadratic causal kernel attention.

\begin{align*}
  (L \circ QK^\top) \cdot V = (L \circ CB^\top) \cdot X
\end{align*}

\paragraph{Rotary Position Embedding.}
Position information is very important in language modeling, and there are three mainstream relative positional encodings: convolution, recursive, and inner product. 
Rotary Position Embedding~\citep{su2021roformer} adds absolute position information $m$ and $n$ to the $Q$ and $K$ matrices in the Quadratic Causal Self-Attention, and calculates the inner product of $QK^\top$ to obtain the relative position encoding matrix.

\begin{align*}
    <f(Q, m), f(K, n)> &= g(Q, K, m - n)
\end{align*}

\paragraph{Shared Expert Isolation.}
The sparse activation mixture of experts architecture aims to train a larger model in fewer training steps with limited computational resources, which often performs better than training a smaller model in more steps.
In the routing expert strategy, to ensure that the experts learn non-redundant general knowledge, Shared Expert Isolation~\citep{dai2024deepseekmoe} shares knowledge by isolating k experts $e(x)$, adding the entire sequence state of the isolated experts to the state of each token of the routing expert.

\begin{align*}
    e(x) &= \sum_{i=1}^{k} e_i(x_i) + \sum_{i=1}^{n - k} e_i(x)
\end{align*}

\paragraph{Parameter Efficient Expert Retrieval.}
In knowledge-intensive modeling tasks, the finer the granularity of the sparse activation mixture of experts, the lower the model perplexity, but the retrieval time of the routing expert strategy will also increase significantly.
Mixture of A Million Experts~\citep{he2024moame} proposes parameter-efficient expert retrieval, which maintains computational efficiency with a large number of experts.

\begin{align*}
    e_i(x) &= \sigma(d_i^\top x) u_i
\end{align*}

\section{Methods}
\label{sec:methods}

Wonderful Matrices is a foundation architecture designed to build efficient and effective models.

\paragraph{Rotary Position Embedding for Hybrid Algorithms.}
First, we prove the availability of Rotary Position Embedding in the hybrid State Space Duality and Quadratic Causal Self-Attention algorithms. Ensuring that the position encoding is consistent for the hybrid algorithm, whether it is training or inference. The method is described in Section \ref{sec:methods:rope_for_hybrid_algorithms}.

\paragraph{Dynamic Mask Attention.}
Second, we propose Dynamic Mask Attention with the same selectivity as State Space Duality. Ensuring that Quadratic Causal Self-Attention can selectively filter past states related to the current state, directly masking irrelevant states in the attention score matrix. The method is described in Section \ref{sec:methods:dma}.

\paragraph{Cross Domain Mixture of Experts.}
Third, we design Cross Domain Mixture of Experts composed of Embedding layers and Linear layers. Ensuring that in dense knowledge tasks such as language modeling, parameters can be fully utilized to store general knowledge and domain-specific knowledge. The method is described in Section \ref{sec:methods:cdmoe}.

\paragraph{Architecture Design.}
Finally, we combine Rotary Position Embedding, State Space Duality, Dynamic Mask Attention, Cross Domain Mixture of Experts to design the Wonderful Matrices architecture in the language modeling task. The architecture is described in Section \ref{sec:methods:architecture}.

\subsection{Rotary Position Embedding for Hybrid Algorithms}
\label{sec:methods:rope_for_hybrid_algorithms}

For example, in the Self-Attention $QK^\top$, the dot product of two vectors $Q_m \cdot K_n$ is calculated, and the result is a scalar, which represents the correlation between position $m$ and position $n$. 
The basic idea of rotary position embedding is to encode the position information as a complex rotary matrix, whose angle is determined by the position index. 
When $QK$ or $CB$ is applied with the Rotary Position Embedding, if an element position is close to the front, its rotation will affect the direction of the $K$ or $B$ vector multiplied with it, thereby affecting the result of the inner product.

The first step is to define the absolute position information embedding function~\ref{eq:rope_for_attn_ssd:abs_pos}. We define four functions to add absolute position information, where the $m$-th position of the $Q$ and $C$ matrices adds absolute position information $m$, and the $n$-th position of the $K$ and $B$ matrices adds absolute position information $n$. Where $\theta$ is the rotation angle.

\begin{align}
    f(q, m) = q e^{i m \theta} \quad 
    f(k, n) = k e^{i n \theta} \quad
    f(c, m) = c e^{i m \theta} \quad 
    f(b, n) = b e^{i n \theta}
    \label{eq:rope_for_attn_ssd:abs_pos}
\end{align}

The second step is to define the score algorithm of the rotary position encoding for hybrid algorithms~\ref{eq:rope_for_attn_ssd:score}. In Appendix~\ref{sec:rope_for_ssd}, we prove how to achieve rotary position encoding in the semiseparable matrix used in State Space Duality. Then we can use the same method to calculate the score with relative position information for each position of $QK$ or $CB$. Where $R_{\Theta}^{d}$ is the rotation matrix, $\Theta$ is the rotation angle set.

\begin{align}
    attn_{score} = f(q, m) R_{\Theta, m - n}^{d} f(k, n)^\top \quad ssd_{score} = f(c, m) R_{\Theta, m - n}^{d} f(b, n)^\top
    \label{eq:rope_for_attn_ssd:score}
\end{align}

The final step is to apply the causal mask $L$ to the score matrix after position encoding, and output the score matrix with position information to $V$ and $X$~\ref{eq:rope_for_attn_ssd:apply_mask} to extract features.

\begin{align}
    y = Attn_{score} \circ L \cdot V \quad y = SSD_{score} \circ L \cdot X
    \label{eq:rope_for_attn_ssd:apply_mask}
\end{align}

In addition, another important reason for unifying the position encoding of Quadratic Causal Self-Attention and State Space Duality into Rotary Position Embedding is to achieve effective position information for linear fast inference~\ref{eq:rope_for_attn_ssd:linear_inference}.
Rotary Position Embedding is a way to achieve relative position encoding with absolute position encoding, without operating the score matrix, it is a relative position encoding method that can be directly used for linear attention, and linear attention can use cache to achieve fast generation. State Space Duality can achieve two generation update methods, one is to use cache, and the other is to directly recursively all previous sequence states to the current sequence state, both of which can use Rotary Position Embedding to achieve relative position encoding. Where $\phi$ and $\varphi$ are non-negative functions.

\begin{align}
    Attn(Q, K, V)_i = \frac{\sum_{j=1}^{n} [R_i \phi(q_i)] [R_j \varphi(k_j)]^\top v_j}{\sum_{j=1}^{n} \phi(q_i) \varphi(k_j)^\top} \quad
    SSD(C, B, X)_i = \frac{\sum_{j=1}^{n} [R_i \phi(c_i)] [R_j \varphi(b_j)]^\top x_j}{\sum_{j=1}^{n} \phi(c_i) \varphi(b_j)^\top}
    \label{eq:rope_for_attn_ssd:linear_inference}
\end{align}

This unified Rotary Position Embedding method allows different sequence transformation algorithms to share the same position information, ensuring the consistency of sequence transformation, and does not need to operate the score matrix directly, and can be used in linear generation. The algorithm matrix of the rotary position embedding is shown in Figure~\ref{fig:rope_for_attn_ssd}. In Appendix~\ref{sec:implementation_code:rope}, an implementation code example of RoPE and its application in Attn and SSD are provided.

\begin{figure}[!t]
    \centering
    \includegraphics[width=\linewidth]{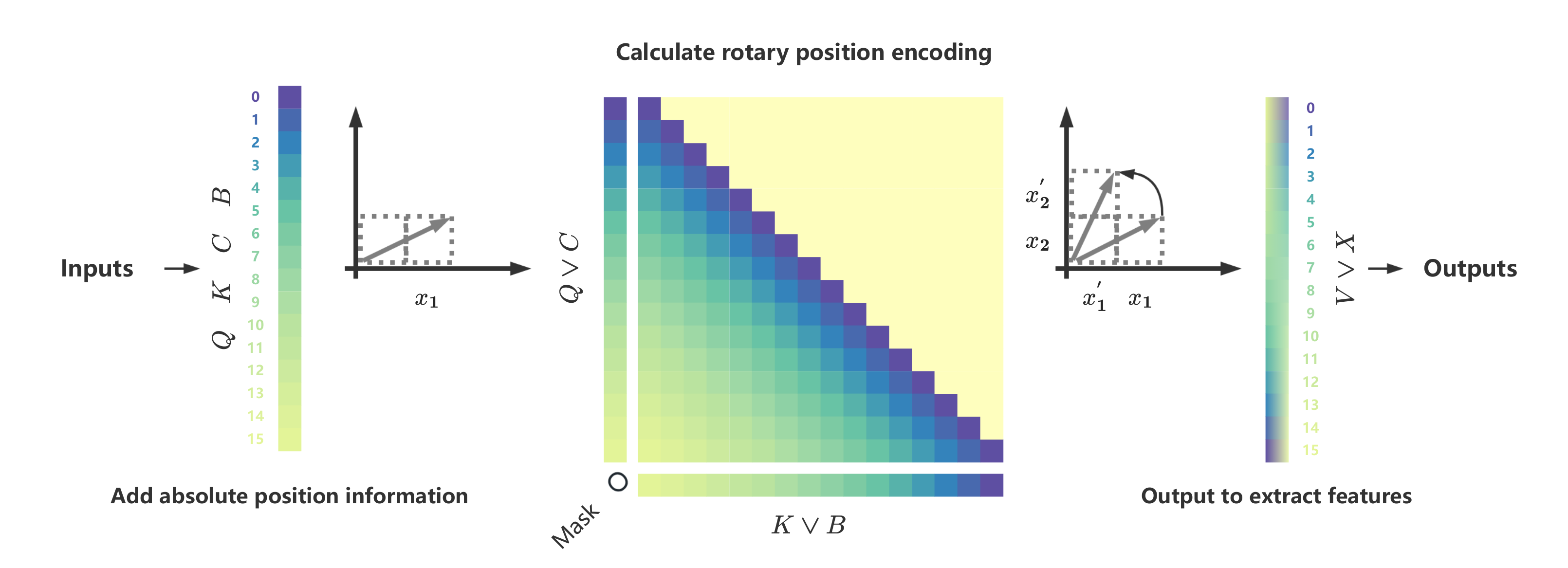}
    \caption{
        \textbf{Rotary Position Embedding}.
        Shows the algorithm of Rotary Position Embedding. In the case of input containing sequence dimension and hidden dimension, first add absolute position information $m$ to the $Q$ and $C$ matrices, add absolute position information $n$ to the $K$ and $B$ matrices, then multiply the rotation matrix $\mathbb{R}_{\Theta, m}^{d}$ and $\mathbb{R}_{\Theta, n}^{d}$ with the $QK$ or $CB$ matrix to obtain the rotary position encoding matrix, and finally apply the mask matrix and output.
    }
    \label{fig:rope_for_attn_ssd}
\end{figure}

\subsection{Dynamic Mask Attention}
\label{sec:methods:dma}

\begin{figure}[!t]
    \centering
    \includegraphics[width=\linewidth]{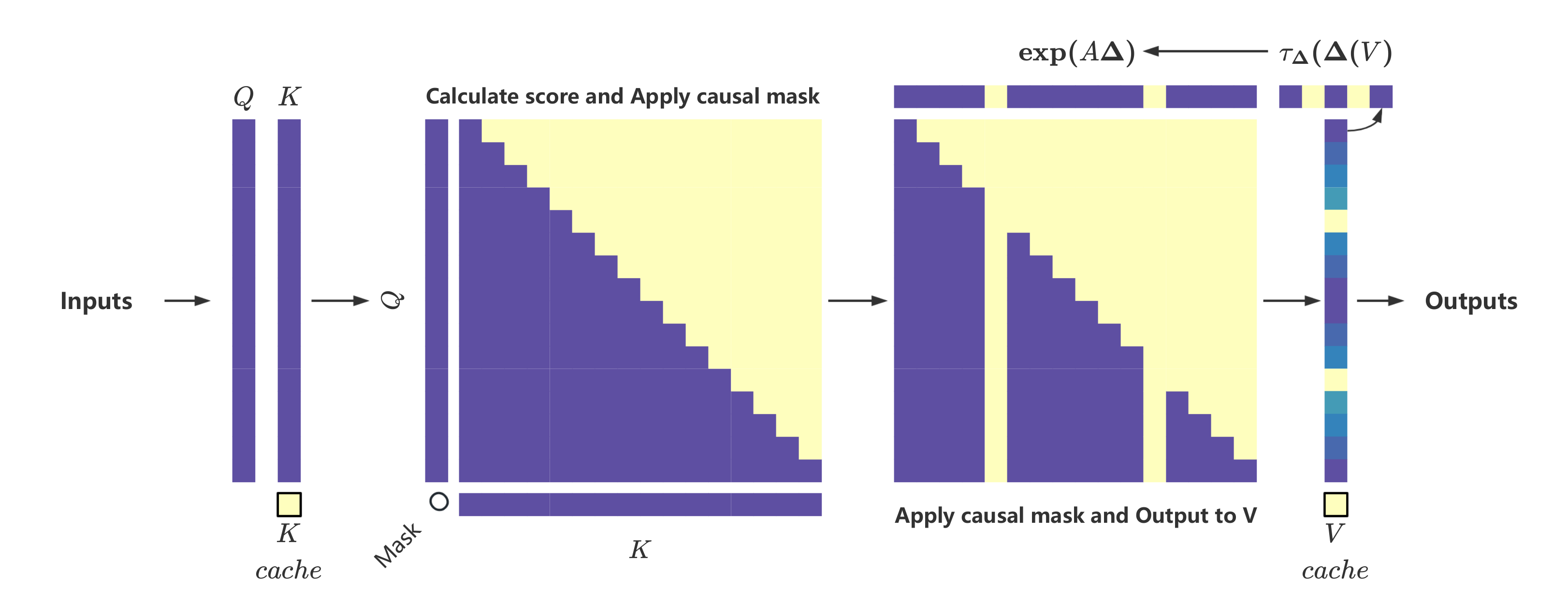}
    \caption{
        \textbf{Dynamic Mask Attention}.
        Shows the algorithm of Dynamic Mask Attention. The input is first projected through the projection function to obtain $QKV$, then the attention score is calculated by the $Q$ state and the concatenated past state $K$ state, the causal mask is applied to the attention score, and finally the score matrix is applied with the dynamic mask related to the concatenated past state $V$ state, and output to the $V$ state.
    }
    \label{fig:dynamic_mask_attn}
\end{figure}

In the Quadratic Causal Self-Attention algorithm, the $Q$ and $K$ related to the entire input sequence are calculated to form the attention score matrix, to extract information from the $V$ related to the input sequence. Linear attention caches the hidden state of $KV$, only calculates the current $Q$ state with the cached $KV$ state, to achieve linear complexity of autoregressive inference. If there is a gate that can selectively filter the information related to the current state from the cached state, then the attention score mask can be dynamically adjusted.

The first step is to define the input-related projection function~\ref{eq:dynamic_mask_attention:input_projection}. We define four functions to project the input state, the projection function of $Q$ $K$ $V$ is the matrix multiplication of the input $x$ with the related matrix weight, while the projection function of the dynamic mask $A$ is to use the zero-order hold technique. Here we explain some of the reasons for using the zero-order hold technique for dynamic masks. First, considering that the states usually contain continuous correlation in language modeling, to reduce the computational cost, we will retain its value every time we receive a continuous correlated state until we receive a new non-continuous correlated state. Second, to maintain this value over time, we introduce a learnable parameter $W_{\Delta}$, which linearly projects the $V$ state to directly obtain the correlation with the past cached state and the current input state. Third, we pass the time step through a non-negative function $\tau_{\Delta}$, because we cannot guarantee that the value domain of the learnable parameter $A$ is non-negative. Finally, we calculate the time step and parameter $A$ through the exponential function to achieve the zero-order hold technique.

\begin{align}
    f(x, W_{Q}) = x W_{Q} \quad f(x, W_{K}) = x W_{K} \quad f(x, W_{V}) = x W_{V} \quad f(V_{cache}, V, W_{\Delta}, A) = \exp(\tau_{\Delta}(\operatorname{concat}(V_{cache}, V) W_{\Delta}) A)
    \label{eq:dynamic_mask_attention:input_projection}
\end{align}

The second step is to calculate the attention score and apply the causal mask~\ref{eq:dynamic_mask_attention:score}. We first concatenate the current $K$ state with the past $K$ state, then perform dot product calculation with the current $Q$ state to obtain the attention score. Finally, apply the causal mask $L$ to the attention score to obtain the causal mask attention score matrix $M$.

\begin{align}
    M = f(x, W_{Q}) \cdot \operatorname{concat}(K_{cache}, f(x, W_{K}))^\top \circ L
    \label{eq:dynamic_mask_attention:score}
\end{align}

The final step is to apply the dynamic mask to the score matrix and output it to the value state~\ref{eq:dynamic_mask_attention:apply_dynamic_mask}. We first concatenate the past $V$ state with the current $V$ state, then perform zero-order hold operation with the time step weight $W_{\Delta}$ and parameter $A$ to obtain the dynamic mask, and finally apply the dynamic mask to the score matrix and perform matrix multiplication with the $V$ state to obtain the final state. Here we explain some additional operations on how the dynamic mask is applied to the score matrix. First, the time-sampled state will be operated by a non-negative function, only the parameter $A$ will have negative values during continuous learning. Second, after the $\Delta A$ state is operated by the exponential function, the negative values will be converted to values less than 1. Finally, if the causal mask is applied to the attention score using addition, then its value domain will be $(-\infty, 0]$, we can convert the part of the dynamic mask less than 1 to $-\infty$ and apply it to the attention score. If the causal mask is applied to the attention score using multiplication, then its value domain will be $[0, 1]$, we can directly perform element-wise multiplication between the dynamic mask and the causal mask, the part originally filled with 0 will not change, while the other parts can be dynamically attenuated or enhanced.

\begin{align}
    y = M \circ f(V_{cache}, V, W_{\Delta}, A) \cdot \operatorname{concat}(V_{cache}, f(x, W_{V}))
    \label{eq:dynamic_mask_attention:apply_dynamic_mask}
\end{align}

This Quadratic Causal Self-Attention algorithm with dynamic mask can selectively filter the information related to the current state from the final superimposed state of self-attention, which can directly mask invalid states, and can attenuate or enhance some states, without causing decay or bias in past states. The algorithm matrix of dynamic mask attention is shown in Figure~\ref{fig:dynamic_mask_attn}. An implementation code example of dynamic mask attention is provided in Appendix~\ref{sec:implementation_code:dynamicmaskattn}.

\subsection{Cross Domain Mixture of Experts}
\label{sec:methods:cdmoe}

\begin{figure}[!t]
    \centering
    \includegraphics[width=\linewidth]{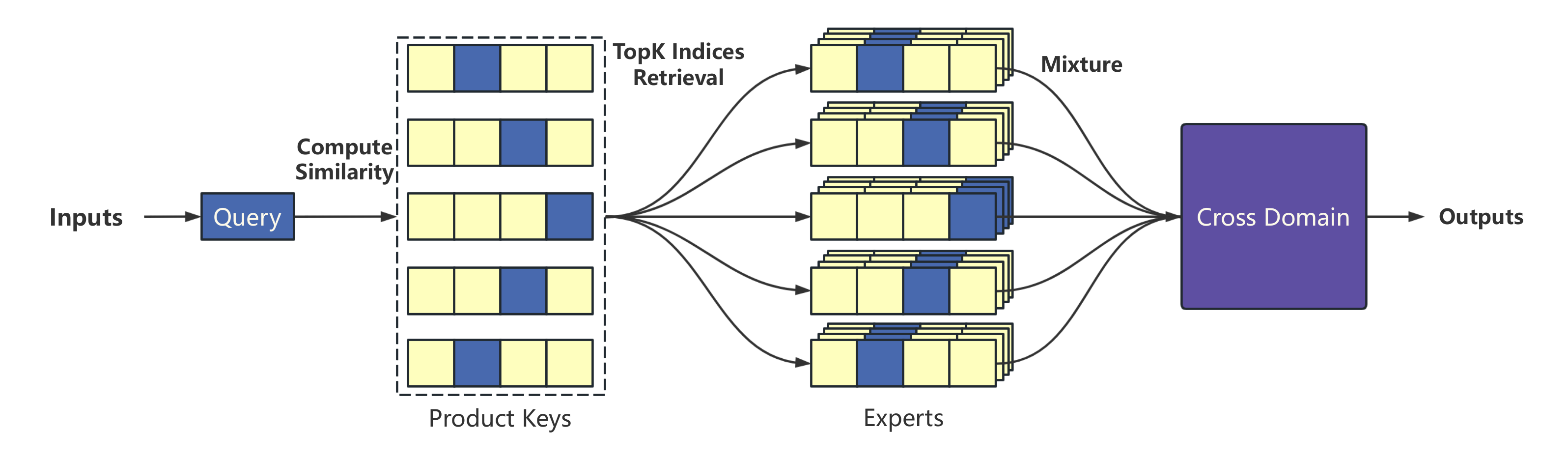}
    \caption{
        \textbf{Cross Domain Mixture of Experts}.
        Shows the algorithm of Cross Domain Mixture of Experts.
        The inputs first passes through the query projection, then calculates the dot product with the keys to obtain the affinity with the private experts, then activates the top K private expert parameters with high affinity, and finally mixes with the cross domain parameters and outputs.
    }
    \label{fig:cdmoe}
\end{figure}

In the conventional mixture of experts strategy, the tokens assigned to different experts need to have common knowledge, so each expert may have redundant parameters for storing the same information. The proportion of expert parameter redundancy increases with the increase in expert granularity and the decrease in the number of expert activations. In Parameter Efficient Expert Retrieval, due to the high granularity of the experts, even if the expert weight rows are all shared, the proportion of expert parameter redundancy reaches an astonishing height due to the low number of expert column activations. If the tokens assigned to different experts have passed through the parameters for storing common knowledge, the parameter redundancy can be reduced.

The first step is to initialize all weights~\ref{eq:cdmoe:init_weights}.
We define query projection matrix weight $W_{Q}$, learnable key parameter $\theta_{K}$, two Embedding matrices $W_{down}$ and $W_{up}$, and two Linear matrices $W_{up}$ and $W_{down}$. Where $\sigma$ represents the activation function, $d_{model}$ represents the model hidden dimension, $d_{cd}$ represents the cross domain dimension, $d_{ret}$ represents the expert retrieval dimension, $n_{e}$ represents the number of experts, and $n_{h}$ represents the number of expert heads.

\begin{align}
    W_{eQ} \in \mathbb{R}^{d_{model} \times n_{h} \times d_{ret}} \quad 
    \theta_{eK} \in \mathbb{R}^{n_{h} \times \sqrt{n_{e}} \times d_{ret}} \quad
    W_{edown}, W_{eup} \in \mathbb{R}^{n_{e} \times d_{model}} \quad
    W_{cdup}, W_{cddown} \in \mathbb{R}^{d_{cd} \times d_{model}}
    \label{eq:cdmoe:init_weights}
\end{align}

The second step is to retrieve the expert state~\ref{eq:cdmoe:retrieve_expert}. 
First, perform matrix multiplication of the input $x$ with $W_{eQ}$ to obtain the query projection, and perform matrix multiplication with $\theta_{eK}^{T}$ to obtain the dot product similarity $g$. 
Then, take the topk expert scores and indices corresponding to each $h$ in the $\sqrt{n_{e}}$ dimension of $g$, and combine them to obtain the similarity score $s$ and expert index $i$.
Finally, perform matrix multiplication of the index position $i$ with $W_{edown}^{T}$ and $W_{eup}^{T}$, which is the embedding extension hidden dimension, to take out the weight rows of the expert dimension, obtaining the embedding states $d$ and $u$ corresponding to the two index positions.

\begin{subequations}
\begin{align}
    g &= x \cdot W_{eQ} \cdot \theta_{eK}^{T} \in \mathbb{R}^{2 \times batch \times seq \times n_{h} \times \sqrt{n_{e}}} \\
    s, i &= \operatorname*{topk}(g, k, dim=-1) \in \mathbb{R}^{batch \times seq \times n_{h} \times k} \\
    d, u &= i \cdot W_{edown}^{T}, i \cdot W_{eup}^{T} \in \mathbb{R}^{batch \times seq \times n_{h} \times k \times d_{model}}
\end{align}
\label{eq:cdmoe:retrieve_expert}
\end{subequations}

The final step is to mix the expert state with the cross domain state~\ref{eq:cdmoe:mix_states}.
First, perform matrix multiplication of the input state $x$ with $d^T$, then perform non-linear activation and multiply with the score $s$ to obtain the expert weight $w$.
The second step is to perform matrix multiplication of the expert weight $w$ with $u$, obtaining $n_{h} \times k$ different expert states, and then summing in the $n{h}, k$ dimension to combine these different expert states to obtain the expert state $\varphi$. 
The third step is to calculate the cross domain state information, perform matrix multiplication of the input $x$ with $W_{cdup}$, then perform non-linear activation and matrix multiplication with $W_{cddown}$ to obtain the cross domain state information $\phi$. 
Finally, add the two states to obtain the final cross domain mixture of experts state $y$.

\begin{subequations}
\begin{align}
    \varphi &= \sum^{n_{h}}_{i=1} \sum^{k}_{j=1} \sigma(x \cdot d_{i, j}^T \times s_{i, j}) \cdot u_{i, j} \in \mathbb{R}^{batch \times seq \times d_{model}} \\
    \phi &= \sigma(xW_{cdup})W_{cddown} \in \mathbb{R}^{batch \times seq \times d_{model}} \\
    y &= \varphi + \phi \in \mathbb{R}^{batch \times seq \times d_{model}}
\end{align}
\label{eq:cdmoe:mix_states}
\end{subequations}

This cross domain with efficient retrieval mixture of experts method not only has a large MLP to store the main state transformation information but also can dynamically combine different small MLPs in the $n_{h}$ dimension. The small MLPs share neurons by aggregating the $h$ singletons retrieved from the shared weight rows, which can efficiently retrieve the expert with the highest affinity for each token and maintain speed as the number of experts increases without rapid decline as in the routing strategy. The algorithm matrix of the cross domain mixture of experts is shown in Figure~\ref{fig:cdmoe}. An implementation code example of CDMoE is provided in Appendix~\ref{sec:implementation_code:cdmoe}.

\subsection{Architecture Design}
\label{sec:methods:architecture}

\begin{figure}[ht]
  \centering
  \includegraphics[width=\linewidth]{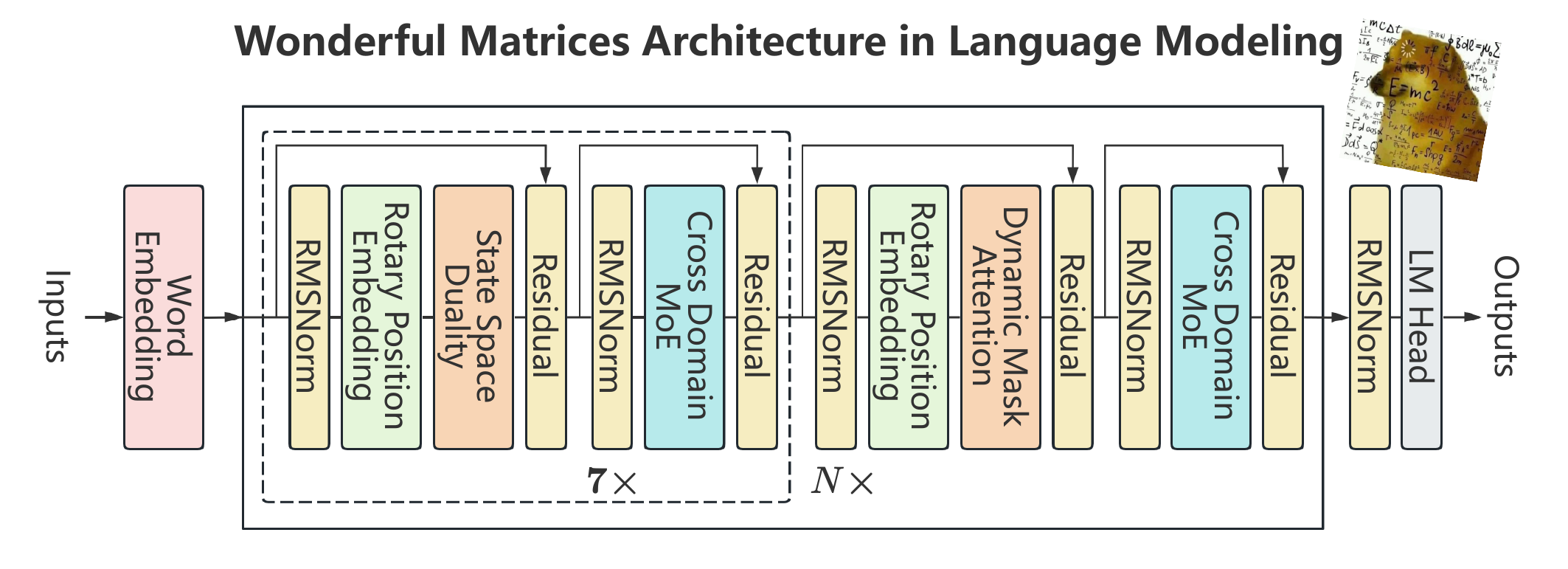}
  \caption{
    \textbf{Wonderful Matrices in Language Modeling: Cheems}.
    Shows the architecture of Wonderful Matrices applied in language modeling, including Word Embedding, RMSNorm, Residual, RoPE, SSD, DMAttn, CDMoE, LM Head modules. The black arrows indicate the calculation order of the modules, the black dashed part indicates stacking this part $7$ times, and the black solid line indicates stacking the entire backbone module part $N$ times. The dog in the upper right corner is the internet-famous Shiba Inu Cheems, which is our sense of humor, allowing us to relax and smile in strict formula derivation work. For the beauty of the table, in subsequent experiments, we will use Cheems as our model name.
  }
  \label{fig:lm_architecture}
\end{figure}

We designed an architecture using these matrices in the language modeling task: Cheems. It first uses Word Embeddings to convert discrete vocabulary into continuous vectors, and outputs the vocabulary probability distribution after passing through Final Norm and LM Head. In the model backbone part, we use RoPE as the position encoding before each sequence transformation module, and use a CDMoE module as the state transformation after the sequence transformation, with input normalization and residual connection between each sequence transformation and state transformation. In the sequence transformation combination method, we stack $7$ SSD modules for each stack, and stack $1$ DMAttn module for each stack, to ensure the performance in-context learning. The architecture of Cheems is shown in Figure~\ref{fig:lm_architecture}.

\section{Empirical Validation}
\label{sec:experiments}

\subsection{Effect of Modules}

\begin{table}[!ht]
    \centering
    \begin{minipage}{.45\linewidth}
        \centering
        \captionsetup{type=table}
        \caption{
        \textbf{$Conv1d + D$ vs. $a_t$ vs. $RoPE$}.
        We use QCAttn to represent Quadratic Causal Self-Attention, SSD to represent State Space Dual, and DMAttn to represent Dynamic Mask Attention. In a single module, QCAttn cannot use $Conv1d + D$ and $a_t$. With the sequence length set to 8192, the perplexity performance of all combinations, $RoPE$ is better than $Conv1d + D$ and $a_t$.
        }
        \label{tab:comparative_pe}
        \begin{tabular}{@{}ccccccccccc@{}}
        \toprule
        \sc{Modules} & \sc{$Conv1d + D$} & \sc{$a_t$} & \sc{$RoPE$} \\
        & \sc{ppl $\downarrow$} & \sc{ppl $\downarrow$} & \sc{ppl $\downarrow$} \\
        \midrule
        QCAttn & --- & --- & 8.38 \\
        SSD & 8.56 & 8.62 & 8.33 \\
        SSD + QCAttn & 8.48 & 8.56 & 8.18 \\
        SSD + DMA(add) & 8.38 & 8.44 & 7.92 \\
        SSD + DMA(mul) & 8.24 & 8.38 & 7.88 \\
        \bottomrule
        \end{tabular}
    \end{minipage}
    \hfill
    \begin{minipage}{.45\linewidth}
        \centering
        \captionsetup{type=table}
        \caption{
            \textbf{MLP vs. CDMoE}.
            We use S to represent the SSD module, A to represent the QCAttn module, M to represent the MLP module, and E to represent the CDMoE module. We strictly construct different models with the same number of parameters, and the perplexity on the pre-training subset gradually decreases as the MoE ratio increases.
        }
        \label{tab:comparative_ffn}
        \begin{tabular}{@{}ccccccccccccccccc@{}}
        \toprule
        \sc{Modules} & \sc{MoE Ratio} & \sc{ppl $\downarrow$} \\
        \midrule
        \makebox[1.25em]{\textbf{SM}} \makebox[1.25em]{\textbf{SM}} \makebox[1.25em]{\textbf{SM}} \makebox[1.25em]{\textbf{SM}} \makebox[1.25em]{\textbf{SM}} \makebox[1.25em]{\textbf{SM}} \makebox[1.25em]{\textbf{SM}} \makebox[1.25em]{\textbf{AM}} & 0\% & 8.18 \\
        \makebox[1.25em]{\textbf{SE}} \makebox[1.25em]{\textbf{SM}} \makebox[1.25em]{\textbf{SM}} \makebox[1.25em]{\textbf{SM}} \makebox[1.25em]{\textbf{SM}} \makebox[1.25em]{\textbf{SM}} \makebox[1.25em]{\textbf{SM}} \makebox[1.25em]{\textbf{AM}} & 6.25\% & 8.06 \\
        \makebox[1.25em]{\textbf{SM}} \makebox[1.25em]{\textbf{SM}} \makebox[1.25em]{\textbf{SM}} \makebox[1.25em]{\textbf{SM}} \makebox[1.25em]{\textbf{SM}} \makebox[1.25em]{\textbf{SM}} \makebox[1.25em]{\textbf{SM}} \makebox[1.25em]{\textbf{AE}} & 6.25\% & 8.12 \\
        \makebox[1.25em]{\textbf{SE}} \makebox[1.25em]{\textbf{SM}} \makebox[1.25em]{\textbf{SM}} \makebox[1.25em]{\textbf{SM}} \makebox[1.25em]{\textbf{SM}} \makebox[1.25em]{\textbf{SM}} \makebox[1.25em]{\textbf{SM}} \makebox[1.25em]{\textbf{AE}} & 12.5\% & 7.96 \\
        \makebox[1.25em]{\textbf{SM}} \makebox[1.25em]{\textbf{SE}} \makebox[1.25em]{\textbf{SE}} \makebox[1.25em]{\textbf{SE}} \makebox[1.25em]{\textbf{SE}} \makebox[1.25em]{\textbf{SE}} \makebox[1.25em]{\textbf{SE}} \makebox[1.25em]{\textbf{AM}} & 37.5\% & 7.52 \\
        \makebox[1.25em]{\textbf{SE}} \makebox[1.25em]{\textbf{SE}} \makebox[1.25em]{\textbf{SE}} \makebox[1.25em]{\textbf{SE}} \makebox[1.25em]{\textbf{SE}} \makebox[1.25em]{\textbf{SE}} \makebox[1.25em]{\textbf{SE}} \makebox[1.25em]{\textbf{AE}} & 50\% & 7.49 \\
        \bottomrule
        \end{tabular} 
    \end{minipage}
\end{table}

\begin{figure}[ht]
    \centering
    \includegraphics[width=\linewidth]{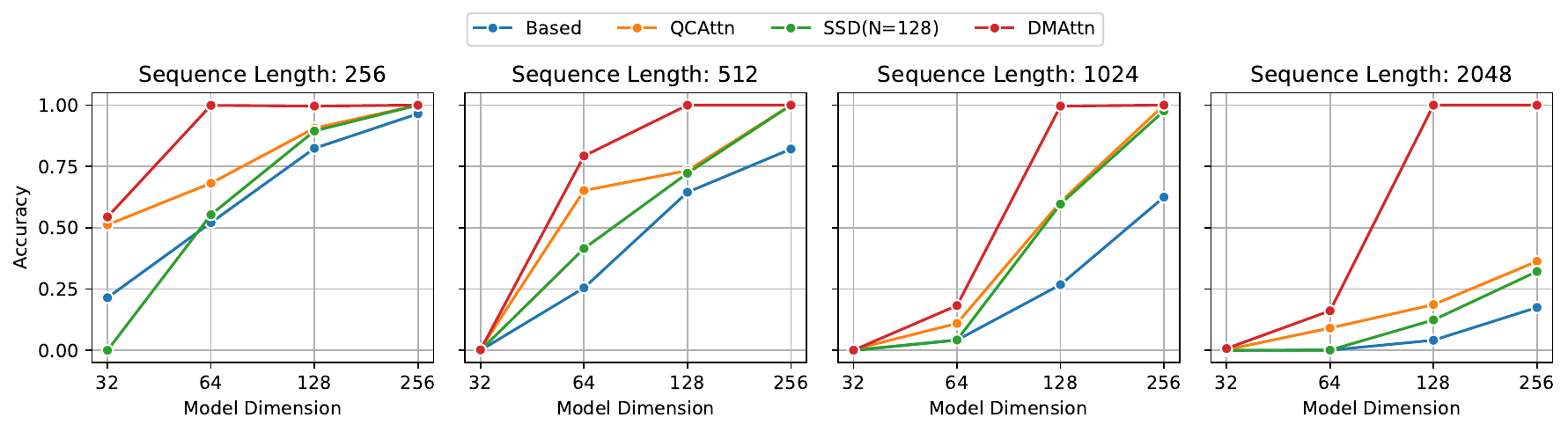}
    \caption{
        \textbf{Multi-Query Associative Recall}.
        We introduced a more difficult version of the original multi-query associative recall task~\citep{arora2024zoology}, including longer sequence lengths, smaller model dimensions, etc. For detailed parameters, see Appendix~\ref{sec:evaluation_parameters:multi_query_associative_recall}. We compared the baseline methods, Quadratic Causal Self-Attention, State Space Dual, and our method, Dynamic Mask Attention, which maintained good effectiveness in most cases.
    }
    \label{fig:mqar}
\end{figure}

\begin{figure}[!h]
    \centering
    \includegraphics[width=\linewidth]{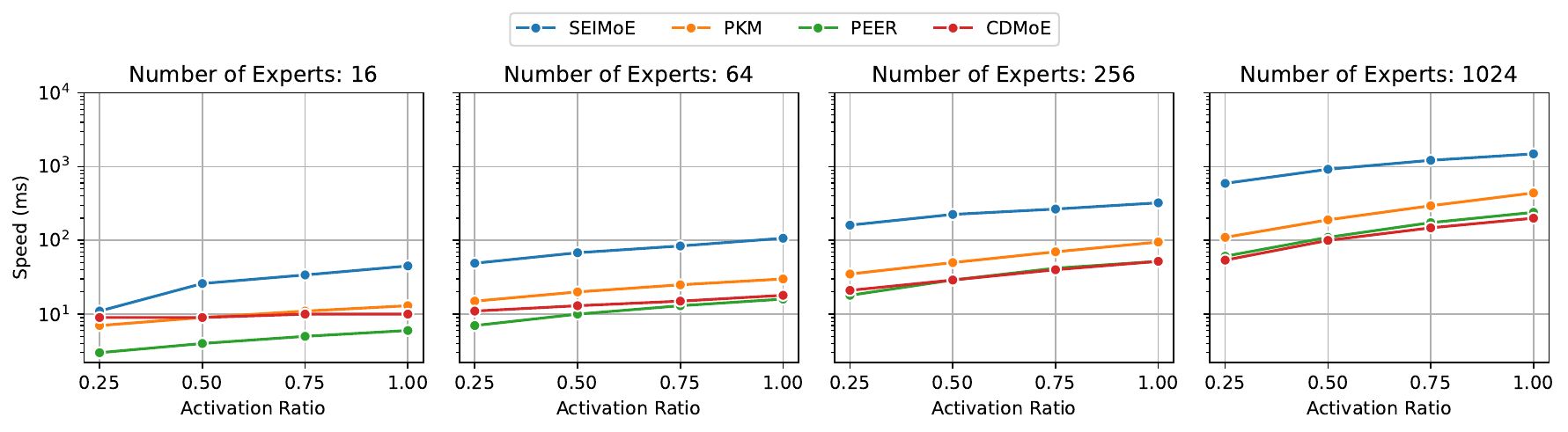}
    \caption{
        \textbf{Multi-Expert Retrieval Mixture}.
        We tested Shared Expert Isolation Mixture of Experts, Product Key Memory~\citep{lample2019pkm}, Parameter Efficient Expert Retrieval and our method in the retrieval mixture speed at different activation ratios $act_{ratio} \times \sqrt{n_{e}}$. When the number of experts reaches $1024$ or more, CDMoE can maintain good efficiency.
    }
    \label{fig:merm}
\end{figure}

\begin{table}[!ht]
    \centering
    \caption{
        \textbf{MoE vs. SEIMoE vs. CDMoE in CEvalBenchmark}.
        We keep the sequence transformation part of the Transformer architecture unchanged, use classic Mixture of Experts, Shared Expert Isolation Mixture of Experts, and our Cross Domain Mixture of Experts as state transformation to construct three models with almost the same total parameters and activation parameters for pre-training. We list the zero-shot and five-shot accuracy of these three models on each subdomain task from CEvalBenchmark~\citep{huang2023ceval}. CDMoE achieved the best results on all tasks.
    }
    \label{tab:comparative_three_moe}
    \begin{tabular}{@{}ccccccccc@{}}
    \toprule
    \sc{Task} & \sc{MoE} & \sc{SEIMoE} & \sc{CDMoE} & \sc{MoE} & \sc{SEIMoE} & \sc{CDMoE} \\
    & \sc{zero-shot $\uparrow$} & \sc{zero-shot $\uparrow$} & \sc{zero-shot $\uparrow$} & \sc{five-shot $\uparrow$} & \sc{five-shot $\uparrow$} & \sc{five-shot $\uparrow$} \\
    \midrule



















    
    STEM & 46.88 & 51.60 & \textbf{53.94} & 47.42 & 52.22 & \textbf{55.01} \\










    
    Social Science & 43.52 & 47.93 & \textbf{48.66} & 41.78 & 46.47 & \textbf{50.64} \\












    Humanities & 46.89 & 51.36 & \textbf{53.20} & 48.54 & 54.10 & \textbf{56.76} \\












    Other & 38.76 & 44.68 & \textbf{48.00} & 41.80 & 48.43 & \textbf{48.94} \\

    Average & 44.29 & 49.29 & \textbf{51.31} & 44.95 & 50.37 & \textbf{52.88} \\
    \bottomrule
    \end{tabular}
\end{table}

In Table~\ref{tab:comparative_pe}, we can see that whether using the State Space Dual algorithm alone or the Quadratic Causal Self-Attention algorithm alone, or using both, the Rotary Position Embedding has the best perplexity performance on long sequences. When the State Space Dual algorithm is combined with Dynamic Mask Attention, not only the position encoding is unified but also the selective state filtering is unified, making this combination perform better on long sequences.

In Table~\ref{tab:comparative_ffn}, we can see that as the proportion of Cross Domain Mixture of Experts using a combination of dense activation Linear layers and sparse activation Embedding layers increases in the entire model, the perplexity on the pre-training subset gradually decreases. However, we also found that in the first and last state transformation modules, the perplexity performance of using completely dense activation MLP modules and using all Cross Domain Mixture of Experts modules is similar. Perhaps this can increase the stability of the input and output of the model backbone. However, for the sake of simple modeling, we chose to use all Cross Domain Mixture of Experts.

In Figure~\ref{fig:mqar}, we can see that in the more difficult multi-query associative recall task, just keeping the dimension of a single attention head at $128$ or above, Dynamic Mask Attention can maintain good effectiveness. Especially when the sequence length is $2048$, the recall ability of the baseline, Quadratic Causal Self-Attention, and State Space Dual are all limited by a large amount of sequence noise, but Dynamic Mask Attention still maintains good associative recall performance.

In Figure~\ref{fig:merm}, we can see that when the number of experts is $16$ or less and the activation ratio $act_{ratio} \times \sqrt{n_{e}}$ is low, using the commonly used Shared Expert Isolation Mixture of Experts is still a good choice. However, once the number of experts reaches $1024$ or more, the retrieval mixture speed of Shared Expert Isolation Mixture of Experts, Product Key Memory, and Parameter Efficient Expert Retrieval will drop sharply, especially Shared Expert Isolation Mixture of Experts. Even if the number of experts is increased while keeping the number of activations unchanged, the retrieval mixture speed will decrease due to the linear cyclic retrieval method. However, Cross Domain Mixture of Experts maintains good retrieval mixture efficiency even with a significant increase in expert granularity.

In Table~\ref{tab:comparative_three_moe}, we can see that by keeping the sequence transformation part of the Transformer architecture unchanged and only using different state transformation parts, that is, the feedforward network. With almost the same total parameters and activation parameters, compared with the classic Mixture of Experts and Shared Expert Isolation Mixture of Experts, our Cross Domain Mixture of Experts achieved the best zero-shot and five-shot accuracy on all subdomain tasks in CEvalBenchmark. The model parameters are similar to the $1.3B$ scale parameter setting in Table~\ref{tab:downstream_evaluation:model}. At the same time, we also have to admire the high quality of the Smollm-Corpus~\citep{benallal2024smollmcorpus} and Chinese Cosmopedia datasets. Compared with using other pre-training datasets, mixing training with them has improved the accuracy of these three models by about $10\%$, especially in the five-shot.

\subsection{Language Modeling}

\begin{figure}[!ht]
    \centering
    \begin{subfigure}{0.48\linewidth}
        \centering
        \includegraphics[width=\linewidth]{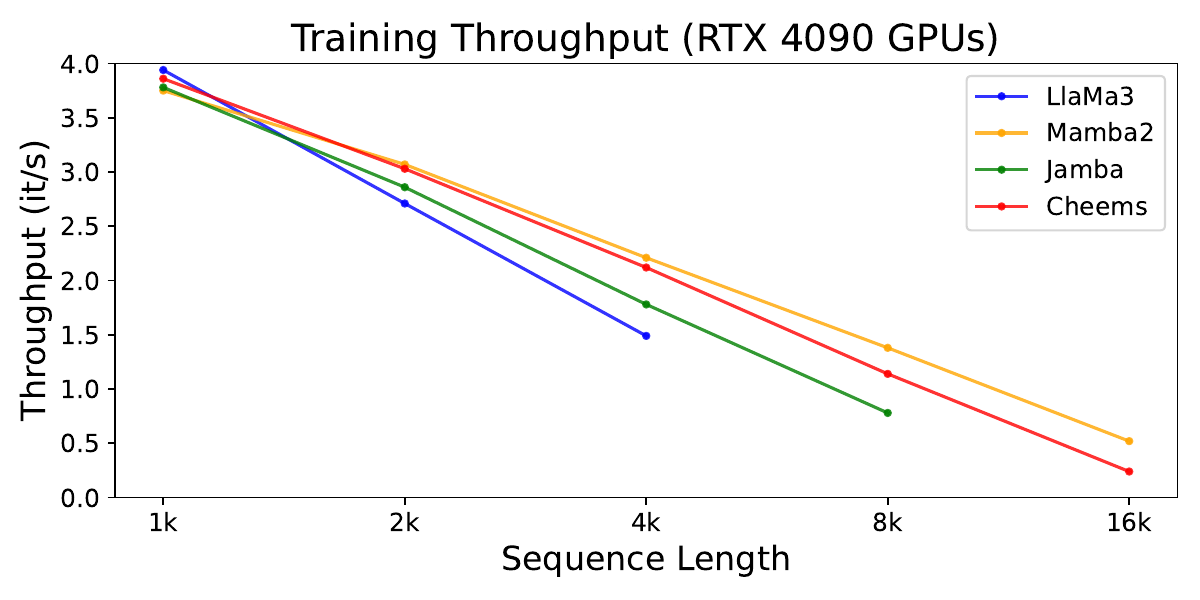}
    \end{subfigure}
    \begin{subfigure}{0.48\linewidth}
        \centering
        \includegraphics[width=\linewidth]{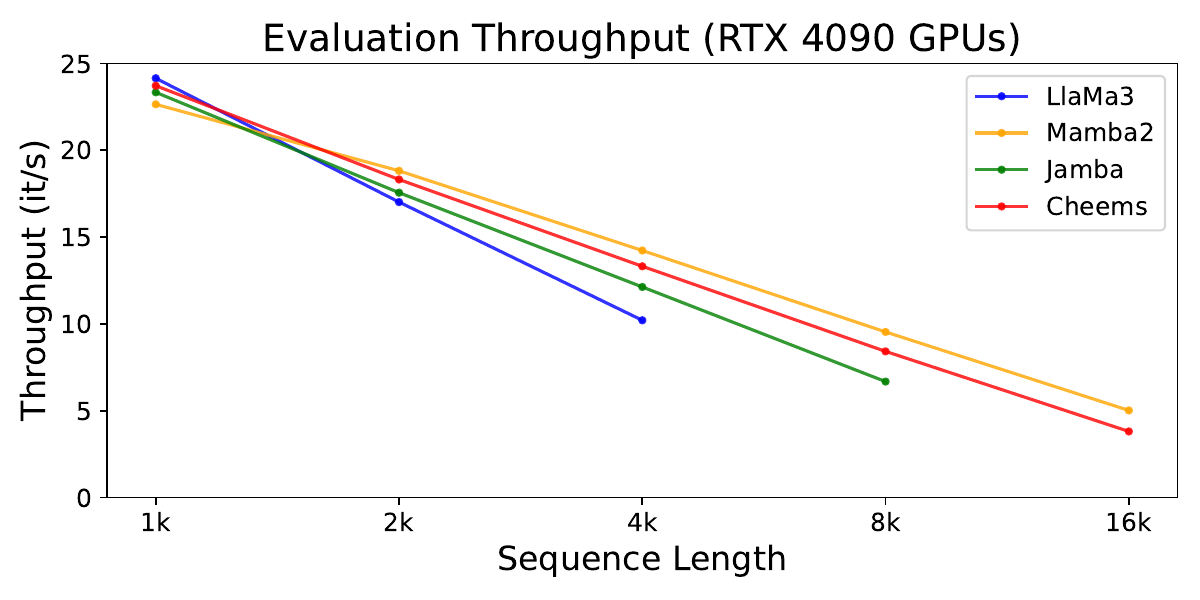}
    \end{subfigure}
    \caption{
        \textbf{Efficient Benchmark}.
        The LlaMa3 architecture that uses QCAttn as the sequence transformation, the Mamba2 architecture that uses SSD as the sequence transformation, the Jamba architecture that uses SSD and QCAttn as the sequence transformation, and the Cheems architecture proposed in this paper. These architectures are train (both forward and backward) and valid (forward only) at different sequence lengths under the 1.3B parameter scale. Cheems is more efficient than LlaMa3 and Jamba, but slightly lower than Mamba2.
    }
    \label{fig:efficient_benchmark}
\end{figure}

\begin{table}[!ht]
    \centering
    \caption{
        \textbf{Effective Benchmark}.
        The LlaMa3 architecture that uses QCAttn as the sequence transformation and SEIMoE as the state transformation, the Mamba2 architecture that uses SSD as the sequence transformation and SEIMoE as the state transformation, the Jamba architecture that uses SSD and QCAttn as the sequence transformation and SEIMoE as the state transformation, and our Cheems.
        The verification results of the models trained under the same conditions. The best results for each parameter scale are shown in bold, followed by underline. For each model parameter scale, Cheems performs better than other models in most cases. We do not provide the perplexity performance of pre-training because the model training completion time is before the gradient accumulation error fix in the Transformers Library~\citep{wolf-etal-2020-transformers}, and there is no reference value for using different gradient accumulation steps before the fix and new experiments after the fix. If our training results are reproduced in the future, the scores on these verification metrics may rise. For the introduction of the verification set and the specific model parameters, see Appendix~\ref{sec:evaluation_parameters:downstream_evaluation}.
    }
    \label{tab:effective_benchmark}
    \begin{tabular}{@{}ccccccccccccc@{}}
    \toprule
    \sc{Model} & \sc{MMLU} & \sc{TriviaQA} & \sc{ARC} & \sc{PIQA} & \sc{HellaSwag} & \sc{OBQA} & \sc{Winogrande} & \sc{Avg} \\
    & \sc{acc $\uparrow$} & \sc{qem $\uparrow$} & \sc{acc $\uparrow$} & \sc{acc $\uparrow$} & \sc{acc $\uparrow$} & \sc{acc $\uparrow$} & \sc{acc $\uparrow$} & \\
    \midrule
    LlaMa3-320M & \underline{33.65} & 8.86 & \textbf{51.68} & 71.42 & 52.30 & \underline{37.02} & 53.15 & 43.99 \\
    Mamba2-320M & 33.10 & \underline{9.36} & 50.72 & 70.24 & 48.62 & 35.16 & 54.17 & 43.07 \\
    Jamba-320M & 33.12 & 9.32 & 50.80 & \underline{71.88} & \underline{52.92} & 36.73 & \underline{55.24} & \underline{44.31} \\
    Cheems-320M & \textbf{34.45} & \textbf{10.38} & \underline{51.57} & \textbf{73.32} & \textbf{53.79} & \textbf{37.42} & \textbf{55.61} & \textbf{45.22} \\
    \midrule
    LlaMa3-1.3B & \underline{37.86} & 20.66 & \textbf{59.82} & 76.05 & 61.65 & \textbf{41.15} & 55.40 & 50.36 \\
    Mamba2-1.3B & 36.28 & 21.28 & 58.02 & 72.26 & 59.48 & 37.98 & 58.72 & 49.07 \\
    Jamba-1.3B & 37.43 & \underline{21.60} & 59.33 & \underline{76.58} & \underline{62.33} & 40.82 & \underline{59.20} & \underline{51.07} \\
    Cheems-1.3B & \textbf{39.08} & \textbf{23.02} & \underline{59.69} & \textbf{78.15} & \textbf{63.63} & \underline{41.12} & \textbf{62.09} & \textbf{52.44} \\
    \bottomrule
    \end{tabular}
\end{table}

We selected LlaMa3~\citep{grattafiori2024llama3herdmodels}, Mamba2~\citep{dao2024ssd}, and Jamba~\citep{lieber2024jamba} under the same conditions as the comparison objects of Cheems. In Figure~\ref{fig:efficient_benchmark}, we can see that the forward and backward propagation efficiency of Cheems has surpassed LlaMa3 and Jamba, and maintains a lower gap with Mamba2. In Table~\ref{tab:effective_benchmark}, we can see that Cheems is better than LlaMa3, Mamba2, and Jamba on most verification metrics. And with the increase of the parameter scale, the performance improvement of Cheems is more obvious.

\section{Discussion}
\begin{figure}[!ht]
    \centering
    \includegraphics[width=\linewidth]{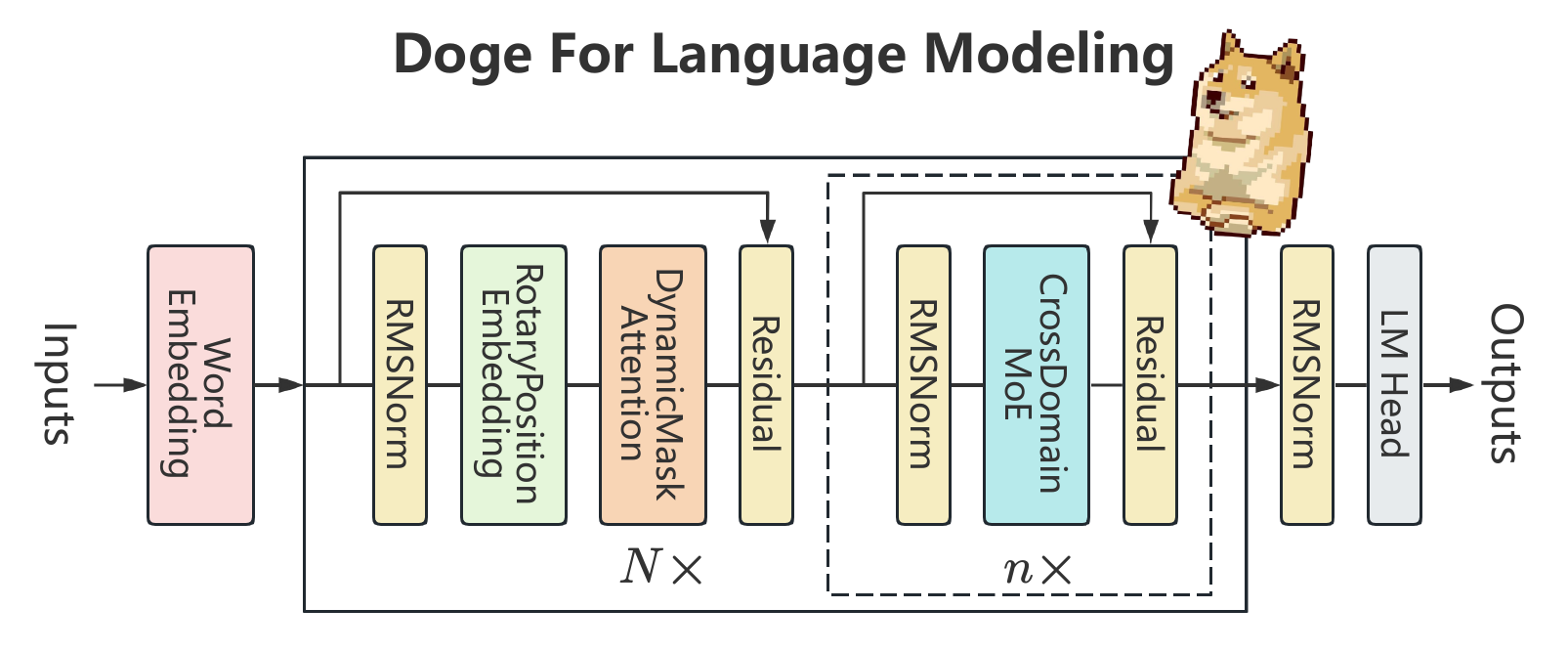}
    \caption{
        \textbf{Doge Architecture}.
        Remove the SSD sequence transformation module in the Cheems architecture and modify the architecture to stack multiple CDMoE state transformation modules after a single DMAttn sequence transformation module. At the same time, Doge can also be understood as a foundation model architecture where Transformer is used during training and SSM is used during inference.
    }
    \label{fig:doge}
\end{figure}

In fact, we encountered many problems when completing this work, including various reasons that caused the mamba-ssm library to not work properly. Before solving this problem, we tried to directly remove the SSD sequence transformation module and modify the architecture to stack multiple MLP or CDMoE state transformation modules after a single DMAttn sequence transformation module as shown in Figure~\ref{fig:doge}. We found that using this model architecture, ensuring that the parameter amount is equal to or less than other architectures for language modeling, there is no significant decrease in most verification metrics. We speculate that on the one hand, DMAttn allows Transformer and SSM to transform each other, and on the other hand, in the current Transformer architecture, there may be some redundancy in the Attn layer. Studying the impact of attention scores on the depth of the layer may be a research direction. We will continue to explore this in future work and will open-source a series of related weights on \url{https://huggingface.co/JingzeShi}.

\section{Conclusion}
\label{sec:conclusion}

This paper explores the idea of modeling by integrating the state space dual algorithm with the quadratic causal self-attention algorithm. We studied the unified position encoding under the hybrid algorithm, proposed dynamic mask attention that can selectively filter information related to the current state, and designed cross domain mixture of experts to reduce parameter redundancy. Finally, this paper verifies that these algorithms achieve advanced performance in language modeling, promoting the development of language modeling in a more efficient and effective direction.

\subsubsection*{Acknowledgments}
We thank our families for their understanding and support in completing this work as independent researchers. At the same time, we also thank Professor Albert Gu of Carnegie Mellon University for providing us with an endorsement of ArXiv, allowing us to engage in scientific research at the undergraduate level.

\printbibliography

\newpage

\appendix

\onecolumn
  
\section{RoPE for SSD}
\label{sec:rope_for_ssd}

\begin{proof}[Proof of equation~\ref{eq:rope_for_attn_ssd:score}]

by definition, $h_0 = B_0 x_0$.
By induction,

\begin{align*}
    h_t &= A_t \dots A_1 B_0 x_0 + A_t \dots A_2 B_1 x_1 + \dots + A_t A_{t-1} B_{t-2} x_{t-2} + A_t B_{t-1} x_{t-1} + B_t x_t
    \\&= \sum_{s=0}^t A_{t:s}^\times B_s x_s
\end{align*}

Multiplying by $C_t$ to produce $y_t$, and vectorizing the equation to $t \in [\mathtt{T}]$ ($\mathtt{T}$ is the sequence length), we derive the matrix transformation form of SSD.

\begin{align*}
    y_t &= \sum_{s=0}^t C_t^{\top} A_{t:s}^\times B_s x_s
    \\
    y &= \mathsf{SSD}(A, B, C)(x) = Mx
    \\
    M_{ji} &\coloneqq C_j^{\top} A_{j} \cdots A_{i+1} B_{i}
\end{align*}

Then the matrix form of SSD is represented using SSS (Sequentially Semiseparable) as $M = \mathsf{SSS}(A, B, C)$, where $M_{ji} = C_j^{\top} A_{j:i} B_i$, and then considering $A$ is just a scalar, rearranged as

\begin{align*}
    M_{ji} = A_{j:i} \cdot (C_j^{\top}B_i)
\end{align*}

Vectorized as

\begin{align*}
    L &\coloneqq \mathsf{1SS}(a) \\
    M &= L \circ (C B^{\top}) \\
\end{align*}

Finally, it is proved that the matrix transformation form of SSD is equivalent to Attention $(L \circ QK^\top) \cdot V = (L \circ CB^\top) \cdot X$.

Now we have enough theoretical support to give rotational positional encoding to the $C$ and $B$ matrices in SSD.

\begin{subequations}
    \begin{align*}
        C_{m} &= f_{C} (x_{m}, m) \\
        B_{n} &= f_{B} (x_{n}, n)
    \end{align*}
\end{subequations}

$C_{m}$ represents the output weight matrix of the $m$-th token corresponding to the word vector $x_{m}$ integrated with the position information $m$,
$B_{n}$ represents the input weight matrix of the $n$-th token corresponding to the word vector $x_{n}$ integrated with the position information $n$.

To utilize the relative positional information between tokens,
we assume that the inner product operation between the $C_{m}$ vector and the $B_{n}$ vector can be represented by a function $g$,
where the input of the function $g$ is the word embedding vectors $x_{m}$ and $x_{n}$,
and their relative positional information $m - n$,
the inner product of $C_{m}$ and $B_{n}$ and their relative positional information $m - n$ is defined as

\begin{equation*}
    <f_{C}(x_{m}, m), f_{B}(x_{n}, n)> = g(x_{m}, x_{n}, m - n)
\end{equation*}

Now, assuming the word embedding vector dimension is $d = 2$,
we have $f_{C}(x_{m}, n) = (W_{C} x_{m})e^{i m \theta}$,
for the first half of the formula $W_{C} x_{m}$,
we know that $W_{C}$ is a two-dimensional matrix,
$x_{m}$ is a two-dimensional vector,
the result of the multiplication is naturally a two-dimensional vector,
represented by $C_{m}$

\begin{align*}
    C_{m} &= 
        \begin{bmatrix} 
            C_{m}^{(1)} \\
            C_{m}^{(2)}
        \end{bmatrix}
    = W_{C} x_{m} = 
        \begin{bmatrix} 
            W_{C}^{(11)} & W_{C}^{(12)} \\
            W_{C}^{(21)} & W_{C}^{(22)}
        \end{bmatrix}
        \begin{bmatrix} 
            x_{m}^{(1)} \\
            x_{m}^{(2)}
        \end{bmatrix}
\end{align*}

For the second half $e^{i m \theta}$,
according to Euler's formula $e^{i x} = \cos(x) + i \sin(x)$,
we have

\begin{align*}
    e^{i m \theta} &= \cos(m \theta) + i \sin(m \theta)
\end{align*}

We know

\begin{align*}
    f_{C}(x_{m}, m) &= (W_{C} x_{m})e^{i m \theta} = C_{m}e^{i m \theta}
\end{align*}

$C_{m}$ is represented in complex form,

\begin{align*}
    C_{m} &=
        \begin{bmatrix} 
            C_{m}^{(1)}, C_{m}^{(2)}
        \end{bmatrix}
    = 
        \begin{bmatrix} 
            C_{m}^{(1)} + i C_{m}^{(2)}
        \end{bmatrix}
\end{align*}

Thus,

\begin{align*}
    f_{C}(x_{m}, m) &= C_{m}e^{i m \theta} = 
        \begin{bmatrix} 
            C_{m}^{(1)} + i C_{m}^{(2)}
        \end{bmatrix} e^{i m \theta}
\end{align*}

According to the above derivation,
we know that $f_{C}(x_{m}, m)$ is the product of two complex numbers,

\begin{align*}
    f_{C}(x_{m}, m) &= C_{m}e^{i m \theta} = 
        \begin{bmatrix} 
            C_{m}^{(1)} + i C_{m}^{(2)}
        \end{bmatrix} \times (\cos(m \theta) + i \sin(m \theta))
\end{align*}

Considering the following two formulas about complex numbers

\begin{align*}
    (a + i b) \times (c + i d) &= ac + i bc + i ad + i^2 bd = (ac - bd) + i (bc + ad) \\
    i^2 &= -1
\end{align*}

We have

\begin{align*}
    C_{m}e^{i m \theta} &= 
        \begin{bmatrix} 
            C_{m}^{(1)} + i C_{m}^{(2)}
        \end{bmatrix} \times (\cos(m \theta) + i \sin(m \theta)) = 
        \begin{bmatrix} 
            C_{m}^{(1)} \cos(m \theta) - C_{m}^{(2)} \sin(m \theta)
        \end{bmatrix} + i
        \begin{bmatrix} 
            C_{m}^{(2)} \cos(m \theta) + C_{m}^{(1)} \sin(m \theta)
        \end{bmatrix}
\end{align*}

Expressing this result as a real vector,

\begin{align*}
    C_{m}e^{i m \theta} &= 
        \begin{bmatrix} 
            C_{m}^{(1)} \cos(m \theta) - C_{m}^{(2)} \sin(m \theta),
            C_{m}^{(2)} \cos(m \theta) + C_{m}^{(1)} \sin(m \theta)
        \end{bmatrix}
\end{align*}

Therefore, $C_{m}$ multiplied by a rotation matrix is obtained.

\begin{align*}
    f_{C}(x_{m}, m) &= (W_{C} x_{m})e^{i m \theta} = C_{m}e^{i m \theta} \\
    &=
    \begin{bmatrix} 
        C_{m}^{(1)} \cos(m \theta) - C_{m}^{(2)} \sin(m \theta),
        C_{m}^{(2)} \cos(m \theta) + C_{m}^{(1)} \sin(m \theta)
    \end{bmatrix} \\
    &=
    \begin{bmatrix} 
        \cos(m \theta) & -\sin(m \theta) \\
        \sin(m \theta) & \cos(m \theta)
    \end{bmatrix}
    \begin{bmatrix} 
        C_{m}^{(1)} \\
        C_{m}^{(2)}
    \end{bmatrix}
\end{align*}

Similarly, $B_{n}$ vector can be obtained

\begin{align*}
    f_{B}(x_{n}, n) &= (W_{B} x_{n})e^{i n \theta} = B_{n}e^{i n \theta} \\
    &=
    \begin{bmatrix} 
        B_{n}^{(1)} \cos(n \theta) - B_{n}^{(2)} \sin(n \theta),
        B_{n}^{(2)} \cos(n \theta) + B_{n}^{(1)} \sin(n \theta)
    \end{bmatrix} \\
    &=
    \begin{bmatrix} 
        \cos(n \theta) & -\sin(n \theta) \\
        \sin(n \theta) & \cos(n \theta)
    \end{bmatrix}
    \begin{bmatrix} 
        B_{n}^{(1)} \\
        B_{n}^{(2)}
    \end{bmatrix}
\end{align*}

The function $g$ can be represented as

\begin{align*}
    g(x_{m}, x_{n}, m - n) &= R 
        \begin{bmatrix} 
            (W_{C} x_{m})(W_{B} x_{n})^*e^{i (m - n) \theta}
        \end{bmatrix}
\end{align*}

where $R$ represents the real part of the complex number $x$,
$(W_{C} x_{m})(W_{B} x_{n})^*$ represents the conjugate of the product of two complex numbers.
Considering

\begin{align*}
    z &= a + i b \\
    z^* &= a - i b
\end{align*}

we have

\begin{align*}
    W_{C} x_{m} &= C_{m} = C_{m}^{(1)} + i C_{m}^{(2)} \\
    W_{B} x_{n} &= B_{n} = B_{n}^{(1)} + i B_{n}^{(2)} \\
    (W_{B} x_{n})^* &= B_{n}^* = B_{n}^{(1)} - i B_{n}^{(2)} \\
    e^{i (m - n) \theta} &= \cos((m - n) \theta) + i \sin((m - n) \theta)
\end{align*}

We now want to prove that

\begin{align*}
    g(x_{m}, x_{n}, m - n) &= R 
        \begin{bmatrix} 
            (W_{C} x_{m})(W_{B} x_{n})^*e^{i (m - n) \theta}
        \end{bmatrix} \\
    &= R
        \begin{bmatrix} 
            (C_{m}^{(1)} + i C_{m}^{(2)})(B_{n}^{(1)} - i B_{n}^{(2)})(\cos((m - n) \theta) + i \sin((m - n) \theta))
        \end{bmatrix} \\
    &= R
        \begin{bmatrix} 
            ((C_{m}^{(1)}B_{n}^{(1)} + C_{m}^{(2)}B_{n}^{(2)}) + i (C_{m}^{(2)}B_{n}^{(1)} - C_{m}^{(1)}B_{n}^{(2)}))(\cos((m - n) \theta) + i \sin((m - n) \theta))
        \end{bmatrix} \\
    &= (C_{m}^{(1)}B_{n}^{(1)} + C_{m}^{(2)}B_{n}^{(2)})\cos((m - n) \theta) - (C_{m}^{(2)}B_{n}^{(1)} - C_{m}^{(1)}B_{n}^{(2)})\sin((m - n) \theta)
\end{align*}

Recalling the vectorized form of SSD,
the $C$ vector at position $m$ and the $B$ vector at position $n$ will perform an inner product operation,
that is,

\begin{align*}
    f_{C}(x_{m}, m) &= 
        \begin{bmatrix} 
            C_{m}^{(1)} \cos(m \theta) - C_{m}^{(2)} \sin(m \theta),
            C_{m}^{(2)} \cos(m \theta) + C_{m}^{(1)} \sin(m \theta)
        \end{bmatrix} \\
    f_{B}(x_{n}, n) &=
        \begin{bmatrix} 
            B_{n}^{(1)} \cos(n \theta) - B_{n}^{(2)} \sin(n \theta),
            B_{n}^{(2)} \cos(n \theta) + B_{n}^{(1)} \sin(n \theta)
        \end{bmatrix}
\end{align*}

We have

\begin{align*}
    <f_{C}(x_{m}, m), f_{B}(x_{n}, n)> &= 
        \begin{bmatrix} 
            C_{m}^{(1)} \cos(m \theta) - C_{m}^{(2)} \sin(m \theta)
        \end{bmatrix}
        \begin{bmatrix} 
            B_{n}^{(1)} \cos(n \theta) - B_{n}^{(2)} \sin(n \theta)
        \end{bmatrix} \\
        &+
        \begin{bmatrix} 
            C_{m}^{(2)} \cos(m \theta) + C_{m}^{(1)} \sin(m \theta)
        \end{bmatrix}
        \begin{bmatrix} 
            B_{n}^{(2)} \cos(n \theta) + B_{n}^{(1)} \sin(n \theta)
        \end{bmatrix} \\
        &= C_{m}^{(1)} \cos(m \theta) B_{n}^{(1)} \cos(n \theta) - C_{m}^{(1)} \cos(m \theta) B_{n}^{(2)} \sin(n \theta) \\
        &- C_{m}^{(2)} \sin(m \theta) B_{n}^{(1)} \cos(n \theta) + C_{m}^{(2)} \sin(m \theta) B_{n}^{(2)} \sin(n \theta) \\
        &+ C_{m}^{(2)} \cos(m \theta) B_{n}^{(2)} \cos(n \theta) + C_{m}^{(2)} \cos(m \theta) B_{n}^{(1)} \sin(n \theta) \\
        &+ C_{m}^{(1)} \sin(m \theta) B_{n}^{(2)} \cos(n \theta) + C_{m}^{(1)} \sin(m \theta) B_{n}^{(1)} \sin(n \theta)
\end{align*}

Considering

\begin{align*}
    \sin(a + b) &= \sin(a)\cos(b) + \cos(a)\sin(b) \\
    \sin(a - b) &= \sin(a)\cos(b) - \cos(a)\sin(b) \\
    \cos(a + b) &= \cos(a)\cos(b) - \sin(a)\sin(b) \\
    \cos(a - b) &= \cos(a)\cos(b) + \sin(a)\sin(b)
\end{align*}

We have

\begin{align*}
    <f_{C}(x_{m}, m), f_{B}(x_{n}, n)> &= 
        C_{m}^{(1)} B_{n}^{(1)} (\cos(m \theta) \cos(n \theta) + \sin(m \theta) \sin(n \theta)) \\
        &+ C_{m}^{(1)} B_{n}^{(2)} (-\cos(m \theta) \sin(n \theta) + \sin(m \theta) \cos(n \theta)) \\
        &+ C_{m}^{(2)} B_{n}^{(1)} (-\sin(m \theta) \cos(n \theta) + \cos(m \theta) \sin(n \theta)) \\
        &+ C_{m}^{(2)} B_{n}^{(2)} (\sin(m \theta) \sin(n \theta) + \cos(m \theta) \cos(n \theta)) \\
        &= C_{m}^{(1)} B_{n}^{(1)} \cos((m - n) \theta) + C_{m}^{(1)} B_{n}^{(2)} \sin((m - n) \theta) \\
        &- C_{m}^{(2)} B_{n}^{(1)} \sin((m - n) \theta) + C_{m}^{(2)} B_{n}^{(2)} \cos((m - n) \theta) \\
        &= (C_{m}^{(1)} B_{n}^{(1)} + C_{m}^{(2)} B_{n}^{(2)})\cos((m - n) \theta) + (C_{m}^{(1)} B_{n}^{(2)} - C_{m}^{(2)} B_{n}^{(1)})\sin((m - n) \theta) \\
        &= (C_{m}^{(1)}B_{n}^{(1)} + C_{m}^{(2)}B_{n}^{(2)})\cos((m - n) \theta) - (C_{m}^{(2)}B_{n}^{(1)} - C_{m}^{(1)}B_{n}^{(2)})\sin((m - n) \theta) \\
        &= g(x_{m}, x_{n}, m - n)
\end{align*}

It is proved that the inner product of the $C$ vector at position $m$ and the $B$ vector at position $n$ is the function $g$.

Finally, using the matrix-vector multiplication form

\begin{align*}
    <f_{C}(x_{m}, m), f_{B}(x_{n}, n)> &= 
        \begin{bmatrix} 
            \begin{bmatrix} 
                \cos(m \theta) & -\sin(m \theta) \\
                \sin(m \theta) & \cos(m \theta)
            \end{bmatrix}
            \begin{bmatrix} 
                C_{m}^{(1)} \\
                C_{m}^{(2)}
            \end{bmatrix}
        \end{bmatrix}^{T}
        \begin{bmatrix} 
            \begin{bmatrix} 
                \cos(n \theta) & -\sin(n \theta) \\
                \sin(n \theta) & \cos(n \theta)
            \end{bmatrix}
            \begin{bmatrix} 
                B_{n}^{(1)} \\
                B_{n}^{(2)}
            \end{bmatrix}
        \end{bmatrix} \\
        &= 
        \begin{bmatrix} 
            C_{m}^{(1)} & C_{m}^{(2)}
        \end{bmatrix}
        \begin{bmatrix} 
            \cos(m \theta) & \sin(m \theta) \\
            -\sin(m \theta) & \cos(m \theta)
        \end{bmatrix}
        \begin{bmatrix} 
            \cos(n \theta) & -\sin(n \theta) \\
            \sin(n \theta) & \cos(n \theta)
        \end{bmatrix}
        \begin{bmatrix} 
            B_{n}^{(1)} \\
            B_{n}^{(2)}
        \end{bmatrix} \\
\end{align*}

Expanding the product of the two rotary matrices, we have

\begin{align*}
    \begin{bmatrix} 
        \cos(m \theta) \cos(n \theta) + \sin(m \theta) \sin(n \theta) & -\cos(m \theta) \sin(n \theta) + \sin(m \theta) \cos(n \theta) \\
        -\sin(m \theta) \cos(n \theta) + \cos(m \theta) \sin(n \theta) & \sin(m \theta) \sin(n \theta) + \cos(m \theta) \cos(n \theta)
    \end{bmatrix}
\end{align*}

Finally, we get

\begin{align*}
    <f_{C}(x_{m}, m), f_{B}(x_{n}, n)> &= 
        \begin{bmatrix} 
            C_{m}^{(1)} & C_{m}^{(2)}
        \end{bmatrix}
        \begin{bmatrix} 
            \cos((m - n) \theta) & -\sin((m - n) \theta) \\
            \sin((m - n) \theta) & \cos((m - n) \theta)
        \end{bmatrix}
        \begin{bmatrix} 
            B_{n}^{(1)} \\
            B_{n}^{(2)}
        \end{bmatrix}
\end{align*}

The above derivation is only for the case of word embedding dimension $d = 2$,
when $d > 2$, the two-dimensional case can be extended to any dimension as follows

\begin{align*}
    f_{\{C, B\}}(x_{m}, m) &= R_{\Theta, m}^{d} W_{\{C, B\}} x_{m}
\end{align*}

The inner product satisfies linearity,
so for any even-dimensional RoPE, we can represent it as a concatenation of the two-dimensional case,
that is, grouping the elements of the word embedding vector in pairs

\begin{align*}
    R_{\Theta, m}^{d} = \begin{bmatrix} 
        \cos m \theta_0 & -sin m \theta_0 & 0 & 0 & \dots & 0 & 0 \\
        \sin m \theta_0 & \cos m \theta_0 & 0 & 0 & \dots & 0 & 0 \\
        0 & 0 & \cos m \theta_1 & -sin m \theta_1 & \dots & 0 & 0 \\
        0 & 0 & \sin m \theta_1 & \cos m \theta_1 & \dots & 0 & 0 \\
        \vdots & \vdots & \vdots & \vdots & \ddots & \vdots & \vdots \\
        0 & 0 & 0 & 0 & \dots & \cos m \theta_{d/2} & -sin m \theta_{d/2-1} \\
        0 & 0 & 0 & 0 & \dots & \sin m \theta_{d/2} & \cos m \theta_{d/2-1} \\
    \end{bmatrix}
\end{align*}

Each group applies the same rotation operation and the rotation angle of each group is calculated as follows:

\begin{align*}
    \Theta &= \{\theta_i = 10000^{-2(i - 1) / d}, i \in [1, 2, \dots, d / 2]\}
\end{align*}
  





\end{proof}

\newpage
\section{Implementation Code}
\label{sec:implementation_code}

\subsection{RoPE}
\label{sec:implementation_code:rope}

\begin{listing}[ht]
    \begin{minted}[breaklines, breakanywhere]{python}
class RotaryEmbedding:
    def __init__(self, dim, max_position_embeddings, base = 10000,scaling_factor = 1.0):
        self.dim, self.base, self.max_position_embeddings, self.scaling_factor = dim, base, max_position_embeddings, scaling_factor
        inv_freq = 1.0 / (self.base ** (torch.arange(0, self.dim, 2) / self.dim))
        self.register_buffer("inv_freq", inv_freq)

    def forward(self, x, position_ids):
        seq_len = torch.max(position_ids) + 1
        if seq_len > self.max_position_embeddings:
            base = self.base * ((self.scaling_factor * seq_len / self.max_position_embeddings) - (self.scaling_factor - 1)) ** (self.dim / (self.dim - 2))
            inv_freq = 1.0 / (base ** (torch.arange(0, self.dim, 2) / self.dim))
        else:
            inv_freq = self.inv_freq
        inv_freq_expanded = inv_freq[None, :, None].expand(position_ids.shape[0], -1, 1)
        position_ids_expanded = position_ids[:, None, :]
        freqs = (inv_freq_expanded @ position_ids_expanded).transpose(1, 2)
        emb = torch.cat((freqs, freqs), dim = -1)
        cos, sin = emb.cos().to(x.dtype), emb.sin().to(x.dtype)
        return cos, sin

def rotate_half(x):
    x1, x2 = x[..., : x.shape[-1] // 2], x[..., x.shape[-1] // 2 :]
    return torch.cat((-x2, x1), dim=-1)

def apply_QK_rotary_pos_emb(q, k, cos, sin, unsqueeze_dim = 2):
    cos, sin = cos.unsqueeze(unsqueeze_dim), sin.unsqueeze(unsqueeze_dim)
    q_embed = (q * cos) + (rotate_half(q) * sin)
    k_embed = (k * cos) + (rotate_half(k) * sin)
    return q_embed, k_embed

def apply_CB_rotary_pos_emb(c, b, cos, sin, unsqueeze_dim = 2):
    cos, sin = cos.unsqueeze(unsqueeze_dim), sin.unsqueeze(unsqueeze_dim)
    c_embed = (c * cos) + (rotate_half(c) * sin)
    b_embed = (b * cos) + (rotate_half(b) * sin)
    return c_embed, b_embed
    \end{minted}
    \caption{PyTorch example of RoPE.}
    \label{lst:rope}
\end{listing}
\clearpage

\subsection{SSD}
\label{sec:implementation_code:ssd}

\begin{listing}[ht]
    \begin{minted}[breaklines, breakanywhere]{python}
def pad_tensor_by_size(input_tensor, pad_size):
    # pad seq_len to be multiple of chunk_len
    return F.pad(input_tensor, (0, 0, 0, 0, 0, pad_size, 0, 0) if len(input_tensor.shape) == 4 else (0, 0, 0, pad_size, 0, 0))

def reshape_into_chunks(input_tensor, pad_size,chunk_len):
    # padding input_tensor with `pad_size` on the seq_len dim (dim=1) and simultaneously splitting it into chunk sequences.
    # b t ... -> b (l c) ...
    if len(pad_tensor_by_size(input_tensor, pad_size).shape) == 3:
        return rearrange(input_tensor, 'b(lc)h->blch', c = chunk_len)
    else:
        return rearrange(input_tensor, 'b(lc)hd->blchd', c = chunk_len)

def segment_sum(input_tensor):
    # uses cumulative sums and masking instead of direct subtractions.
    chunk_len = input_tensor.size(-1)
    # expand input tensor to have an additional dimension and repeat along that dimension
    # [..., chunk_len] -> [..., chunk_len, chunk_len]
    input_tensor = input_tensor[..., None].expand(*input_tensor.size(), chunk_len)
    # create a lower triangular mask with the diagonal set to 0 to 0 out elements above diag
    mask = torch.tril(torch.ones(chunk_len, chunk_len, dtype = torch.bool), diagonal = -1)
    input_tensor = input_tensor.masked_fill(~mask, 0)
    # compute actual cumsum
    tensor_segsum = torch.cumsum(input_tensor, dim=-2)
    # apply mask to keep only the lower triangular part of the cumulative sum result
    mask = torch.tril(torch.ones(chunk_len, chunk_len, dtype = torch.bool), diagonal = 0)
    tensor_segsum = tensor_segsum.masked_fill(~mask, -torch.inf)
    return tensor_segsum
    \end{minted}
    \caption{Example SSD helper function in PyTorch}
    \label{lst:ssd_helper}
\end{listing}

\begin{listing}[ht]
    \begin{minted}[breaklines, breakanywhere]{python}
def ssd(X, dt, A, B, C, chunk_len, D):
    seq_len = X.size(1)
    pad_size = (chunk_len - seq_len % chunk_len) % chunk_len
    D_residual = rearrange(D, '...->...1') * pad_tensor_by_size(X, pad_size)
    # discretize X and A
    X, A = X * rearrange(dt, '...->...1'), A.to(x.dtype) * dt
    # rearrange into blocks/chunks
    X, A, B, C = [reshape_into_chunks(t, pad_size, chunk_len) for t in (X, A, B, C)]
    # compute cumulative sum of A
    A = rearrange(A, 'bclh->bhcl', l = chunk_len)
    A_cumsum = torch.cumsum(A, dim = -1)
    # compute the output for each intra-chunk (diagonal blocks)
    # this is the analog of a causal mask
    L = torch.exp(segment_sum(A))
    # contraction of C and B to get G (attention-weights like)
    G = (rearrange(C, 'blchn->blc1hn') * rearrange(B, 'blchn->bl1chn')).sum(dim = -1) # shape: (b, c, l, s, h)
    # compute M, equivalent to applying attention mask to weights
    M_intermediate = rearrange(G, '...->...1') * rearrange(L, 'bhcst->bcsth1')
    M = M_intermediate.sum(dim = -1)
    # compute Y_diag (apply to values)
    Y_diag = (rearrange(M, '...->...1') * rearrange(X, 'blchp->bl1chp')).sum(3)
    # (right term of low-rank factorization of off-diagonal blocks; B terms)
    decay_states = torch.exp((A_cumsum[:, :, :, -1:] - A_cumsum))
    B_decay_contraction = B * rearrange(decay_states, 'bhcl->bclh1')
    # permute back B * decay states
    states=(rearrange(B_decay_contraction, 'bclhs->bchls1') * rearrange(X, 'blchp->blhc1p')).sum(dim = 3).permute(0, 1, 2, 4, 3)
    previous_states = torch.zeros_like(states[:, :1])
    states = torch.cat([previous_states, states], dim = 1)
    decay_chunk = torch.exp(segment_sum(F.pad(A_cumsum[:, :, :, -1], (1, 0))))
    states_permuted = states.permute(0, 2, 1, 3, 4)
    result = (decay_chunk[..., None, None] * states_permuted[:, :, None, ...]).sum(dim = 2)
    new_states = result.permute(0, 2, 1, 3, 4)
    states=new_states[:, :-1]
    # compute state -> output conversion per chunk
    # (left term of low-rank factorization of off-diagonal blocks; C terms)
    # compute Yoff
    C_times_states = rearrange(C, 'bclhn->bclh1n') * rearrange(states, 'bchpn->bc1hpn')
    Y_off = (C_times_states.sum(-1) * rearrange(torch.exp(A_cumsum), 'bhcl->bclh1'))
    # add output of intra-chunk and inter-chunk terms (diagonal and off-diagonal blocks)
    y = rearrange(Y_diag + Y_off, 'bclhp->b(cl)hp') + D_residual
    # cutting off padded chunks
    if pad_size > 0:
        y=y[:, : seq_len, :, :]
    return y
    \end{minted}
    \caption{Example SSD algorithm in PyTorch. We have changed some slow methods to faster ones. The original SSD algorithm implementation can be found in the Mamba2 paper.}
    \label{lst:ssd_algorithm_pytorch}
\end{listing}

\begin{listing}[ht]
    \begin{minted}[breaklines, breakanywhere]{python}
class SSD:
    def __init__(self, d_model, n_heads, n_groups, d_state, chunk_len, max_position_embedding):
        self.n_heads, self.n_groups, self.d_head, self.d_state, self.chunk_len = n_heads, n_groups, d_model // n_heads, d_state, chunk_len
        # Initialize parameters
        self.C_proj = nn.Linear(d_model, self.n_groups * self.d_state)
        self.B_proj = nn.Linear(d_model, self.n_groups * self.d_state)
        self.A = nn.Parameter(torch.ones(self.n_heads))
        self.dt_proj = nn.Linear(d_model, self.n_heads)
        self.X_proj = nn.Linear(d_model, self.n_heads * self.d_head)
        self.D = nn.Parameter(torch.ones(self.n_heads))
        self.out_proj = nn.Linear(d_model, d_model)
        # Rotary Position Embedding
        self.BC_rotary_emb = RotaryEmbedding(self.d_state, max_position_embedding)

    def forward(self, x):
        """
        Notations: b - batch size d - d_model 
                    h - n_heads p - d_head n - d_state g - n_groups
                    t - target sequence length s - source sequence length
                    c - n_chunks l - chunk_len
        """
        # linear projection C B X
        B = rearrange(self.B_proj(x), 'bt(gn)->btgn', g = self.n_groups, n = self.d_state).repeat(1, 1, self.n_heads // self.n_groups, 1)
        C = rearrange(self.C_proj(x), 'bt(gn)->btgn', g = self.n_groups, n = self.d_state).repeat(1, 1, self.n_heads // self.n_groups, 1)
        X = rearrange(self.X_proj(x),'bt(hp)->bthp',h=self.n_heads)
        # apply rotary position embedding to C and B
        cos, sin = self.BC_rotary_emb(x, position_ids)
        C, B = apply_CB_rotary_pos_emb(C, B, cos, sin)
        dt = F.softplus(self.dt_proj(x))
        if mamba_libray:
            y = mamba_chunk_scan_combined(X, dt, self.A, B, C, self.chunk_len, self.D)
        else:
            y = ssd(X, dt, self.A, B, C, self.chunk_len, self.D)
        y = self.out_proj(rearrange(y, 'bthp->bt(hp)'))
        return y
    \end{minted}
    \caption{Example SSD implementation in PyTorch}
\end{listing}

\clearpage

\subsection{DynamicMaskAttn}
\label{sec:implementation_code:dynamicmaskattn}

\begin{listing}[ht]
    \begin{minted}[breaklines, breakanywhere]{python}
class DMAttn:
    def __init__(self, d_model, n_heads, max_position):
        self.n_heads, self.d_head = n_heads, d_model // n_heads
        # Initialize Parameters
        self.Q_proj = Linear(d_model, d_model)
        self.K_proj = Linear(d_model, d_model)
        self.A = nn.Parameter(torch.ones(n_heads))
        self.dt_proj = Linear(d_model, n_heads)
        self.V_proj = Linear(d_model, d_model)
        self.out_proj = Linear(d_model, d_model)
        # Rotary Position Embedding
        self.QK_rotary_emb = RotaryEmbedding(self.d_head, max_position)

    def forward(self, x, causal_mask, position_ids, past_kv):
        """
        Notation: b - batch t - length d - d_model h - n_heads p - d_head
        """
        # linear projection Q K V
        Q, K, V = self.Q_proj(x), self.K_proj(x), self.V_proj(x)
        # split into multiple heads
        Q = rearrange(Q, "bt(hp)->bhtp", h = self.n_heads)
        K = rearrange(K, "bt(hp)->bhtp", h = self.n_heads)
        V = rearrange(V, "bt(hp)->bhtp", h = self.n_heads)
        # apply rotary position embedding to Q and K
        cos, sin = self.QK_rotary_emb(x, position_ids)
        Q, K = apply_QK_rotary_pos_emb(Q, K, cos, sin)
        # concatenate past key value
        K, V = past_kv.update(K, V)
        # compute attention score matrix
        attn_score = torch.matmul(Q, K.transpose(-2, -1)) / math.sqrt(self.d_head)
        # add mask to attention score
        dt = self.dt_proj(rearrange(V, "bhtp->btd"))
        dynamic_mask = torch.exp(self.A * F.softplus(dt))
        dynamic_mask = rearrange(dynamic_mask, "bth->bht") < 1.0
        mask = causal_mask.masked_fill(dynamic_mask[:, :, None, :], '-inf')
        attn_score = F.softmax(attn_score + mask, dim = -1)
        # apply attention score to V states
        y = torch.matmul(attn_score, V)
        y = self.out_proj(rearrange(y, "bhtp->bt(hp)"))
        return y
    \end{minted}
    \caption{Example Dynamic Mask Attention implementation in PyTorch}
    \label{lst:dynamic_mask_attn}
\end{listing}
\clearpage

\subsection{Cross Domain Mixture of Experts}
\label{sec:implementation_code:cdmoe}

\begin{listing}[ht]
    \begin{minted}[breaklines, breakanywhere]{python}
class CDMoE:
    def __init__(self, act, d_model, d_cd, d_ret, n_experts, n_heads, k_per_head):
        self.act_fn, self.n_heads, self.k_per_head = ACT2FN[act], n_heads, k_per_head
        # Queries and Keys
        self.queries = Linear(d_model, d_ret * n_heads)
        self.num_keys = math.sqrt(n_experts)
        self.keys = Parameter(torch.zeros(n_heads, self.num_keys, 2, d_ret // 2))
        # Experts
        self.down_embed = Embedding(n_experts, d_model)
        self.up_embed = Embedding(n_experts, d_model)
        # Cross Domain
        self.up_proj = Linear(d_model, d_cd)
        self.down_proj = Linear(d_cd, d_model)

    def forward(self, x):
        """
        Notation: b - batch t - length d - d_model n - d_retrieval
                  h - n_heads p - 2 for product key k - number of keys
        """
        # get similarity with queries and keys
        queries = self.queries(x)
        queries = rearrange(queries, 'bt(phn)->pbthn', p = 2, h = self.n_heads)
        # get experts with the highest similarity
        sim = einsum('pbthn,hkpn->pbthk', queries, self.keys)
        (s_x, s_y), (i_x, i_y) = sim.topk(self.k_per_head, dim = -1)
        all_s = einx.add('... i, ... j -> ... (i j)', s_x, s_y)
        all_i = einx.add('... i, ... j -> ... (i j)', i_x * self.num_keys, i_y)
        s, pk_i = all_s.topk(self.k_per_head, dim = -1)
        i = all_i.gather(-1, pk_i)
        down_embed, up_embed = self.down_embed(i), self.up_embed(i)
        # mix experts states with cross domain states
        experts_w = self.act_fn(einsum('btd,bthkd->bthk', x, down_embed) * s)
        experts_states = einsum('bthk,bthkd->btd', experts_w, up_embed)
        cross_domain_states = self.down_proj(self.act_fn(self.up_proj(x)))
        y = cross_domain_states + experts_states
        return y
    \end{minted}
    \caption{Example CDMoE implementation in PyTorch}
    \label{lst:cdmoe}
\end{listing}

\clearpage

\newpage
\section{Evaluation Parameters}
\label{sec:evaluation_parameters}

\subsection{Multi-Query Associative Recall}
\label{sec:evaluation_parameters:multi_query_associative_recall}

\begin{table}[!ht]
    \centering
    \caption{
        \textbf{Data Parameters}.
        We introduce a more challenging task version based on the original multi-query associative recall~\citep{arora2024zoology}, where tokens that are not query/key/value are replaced with random tokens. We also use more key-value pairs and longer sequence lengths. For each sequence length $T \in \{256, 512, 1024, 2048\}$, we use $T/4$ key-value pairs. The total vocabulary size is $8192$, with approximately $250k$ training samples and $1k$ test samples.
    }
    \label{tab:multi_query_associative_recall:data}
    \begin{tabular}{@{}ccccccccccccccccc@{}}
    \toprule
    \sc{vocab} & \sc{seq len} & \sc{kv pairs} & \sc{train examples} & \sc{test examples} & \sc{powar a} & \sc{batch} & \sc{max epochs} \\
    \midrule
    8192 & 256 & 64 & $2^{18}$ & $2^{10}$ & 0.01 & 256 & 64 \\
    8192 & 512 & 128 & $2^{18}$ & $2^{10}$ & 0.01 & 128 & 64 \\
    8192 & 1024 & 256 & $2^{18}$ & $2^{10}$ & 0.01 & 64 & 64 \\
    8192 & 2048 & 512 & $2^{18}$ & $2^{10}$ & 0.01 & 32 & 64 \\
    \bottomrule
    \end{tabular}
\end{table}

\begin{table}[!ht]
    \centering
    \caption{
        \textbf{Model Parameters}.
        These algorithms can all split into multiple heads, so we set them to single heads and use common single head dimensions $d_{model} \in \{32, 64, 128, 256\}$. For fairness, the SSD algorithm is different from the validation structure in Mamba2~\citep{dao2024ssd} and we remove the one-dimensional causal convolution and gated MLP. All algorithms use the structure of sequence transformation to state transformation and stack 2 layers. In preparation for subsequent algorithm mixing, the learning rate for each dimension of these algorithms is the same.
    }
    \label{tab:multi_query_associative_recall:model}
    \begin{tabular}{@{}ccccccccccc@{}}
    \toprule
    \sc{Algorithm} & $d_{model}$ & $n_{layers}$ & $n_{heads}$ & $d_{state}$ & chunk\_len & \sc{leaning rate}\\
    \midrule
    QCAttn & 32/64/128/256 & 2 & 1 & --- & --- & 4e-4/3e-4/2e-4/1e-4 \\
    SSD & 32/64/128/256 & 2 & 1 & 128 & 256 & 4e-4/3e-4/2e-4/1e-4 \\
    DMAttn & 32/64/128/256 & 2 & 1 & --- & --- & 4e-4/3e-4/2e-4/1e-4 \\
    \bottomrule
    \end{tabular}
\end{table}

\begin{figure}[!ht]
    \centering
    \includegraphics[width=\textwidth]{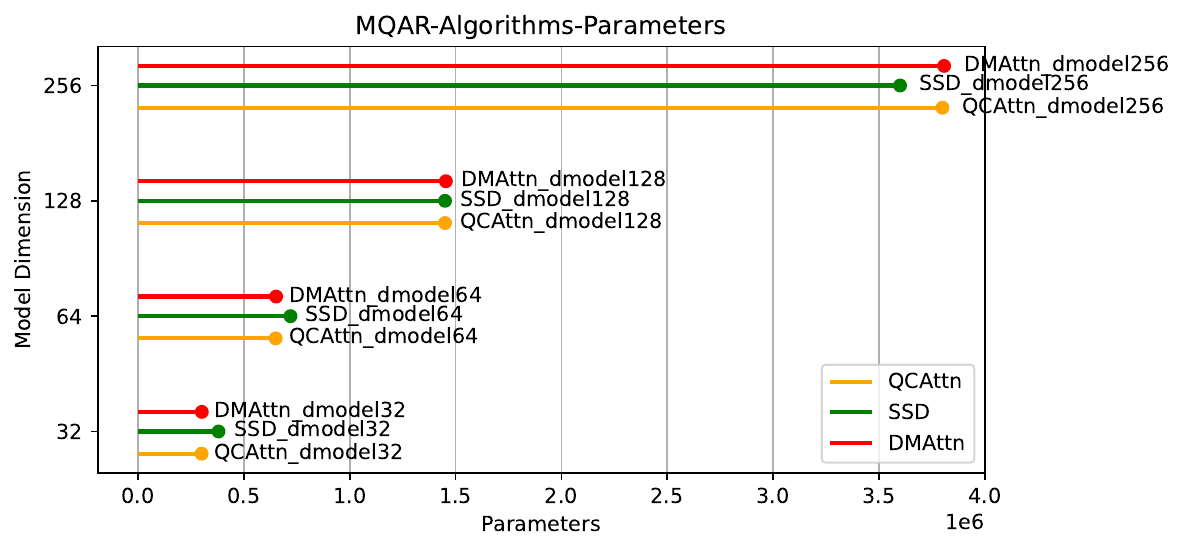}
    \caption{
        \textbf{Different Algorithms Parameters}.
        At different dimensional scales, the number of parameters of DMAttn is not much different from QCAttn. SSD increases the number of parameters less when increasing the dimensional scale.
    }
\end{figure}

\subsection{Downstream Evaluation}
\label{sec:evaluation_parameters:downstream_evaluation}


To avoid score bias in downstream tasks due to different training data, we retrain four model architectures, including Llama3 using the QCAttn algorithm, Mamba2 using the SSD algorithm, Jamba using the hybrid of QCAttn and SSD, and our architecture. We train models of two scales, 360M and 1.3B, with parameters referenced in the table\ref{tab:downstream_evaluation:model}.

\begin{itemize}
    \item All models are trained on the Smollm-Corpus~\citep{benallal2024smollmcorpus} dataset using the NeoX tokenizer.
    \item The training environment is the NVIDIA open-source PyTorch image~\citep{pytorch} version 24.2, which is compatible with the cuda kernel SSD algorithm in the mamba-ssm library.
    \item Training is completed using the Trainer class in the Transformers~\citep{wolf-etal-2020-transformers} library.
    \item AdamW optimizer hyperparameters $\beta_1=0.9, \beta_2=0.999$ and $weight\_decay=0.01$.
    \item The linear warm-up steps are $10\%$ of the total steps, reaching the maximum learning rate of $2e-4$, and then cosine decay to the minimum learning rate of $2e-5$.
    \item No bias terms.
    \item RMSNorm instead of LayerNorm.
    \item Learnable residual connections.
\end{itemize}

For downstream evaluation, we use LM evaluation harness from EleutherAI~\citep{eval-harness}, the validation dataset includes the following tasks:

\begin{itemize}
    \item MMLU~\citep{hendrycks2021measuring}
    \item TriviaQA~\citep{m2017triviaqa}
    \item ARC~\citep{clark2018think}
    \item PIQA~\citep{bisk2020piqa}
    \item HellaSwag~\citep{zellers2019hellaswag}
    \item OBQA~\citep{mihaylov2018can}
    \item Winogrande~\citep{sakaguchi2021winogrande}
\end{itemize}

\begin{table}[!ht]
    \centering
    \caption{
        \textbf{Model Parameters}.
        For fairness, we adjust the important parameters of these four models to be as close in size as possible, and ensure that the total parameters and activation parameters of the four models are as close as possible by adding routing mixture of experts in LlaMa3 and Mamba2, and carefully adjusting the feedforward network expansion size of LlaMa3, Mamba2, and Jamba. Finally, we obtain models of two scales, 320M and 1.3B.
    }
    \label{tab:downstream_evaluation:model}
    \begin{tabular}{@{}ccccccccccc@{}}
    \toprule
    \sc{Model} & $d_{model}$ & $n_{layers}$ & $n_{heads}$ & $d_{state}$ & chunk\_len & $n_{experts}$ & \sc{leaning rate} & \sc{batch size} \\
    \midrule
    LlaMa3-320M & 768 & 24 & 12 & --- & --- & 4 & 3e-4 & 1M tokens \\
    Mamba2-320M & 768 & 24 & 12 & 128 & 256 & 4 & 3e-4 & 1M tokens \\
    Jamba-320M & 768 & 24 & 12 & 128 & 256 & 4 & 3e-4 & 1M tokens \\
    Cheems-320M & 768 & 24 & 12 & 128 & 256 & 3072 & 3e-4 & 1M tokens \\
    \midrule
    LlaMa3-1.3B & 2048 & 24 & 32 & --- & --- & 4 & 2e-4 & 2M tokens \\
    Mamba2-1.3B & 2048 & 24 & 32 & 128 & 256 & 4 & 2e-4 & 2M tokens \\
    Jamba-1.3B & 2048 & 24 & 32 & 128 & 256 & 4 & 2e-4 & 2M tokens \\
    Cheems-1.3B & 2048 & 24 & 32 & 128 & 256 & 8192 & 2e-4 & 2M tokens \\
    \bottomrule
    \end{tabular}
\end{table}
  
\end{document}